\documentclass{article} 
\usepackage{nips13submit_e,times}
\usepackage{hyperref}
\usepackage{url}
\usepackage{amsfonts}
\usepackage{amsmath}
\usepackage{algorithm}
\usepackage{algorithmicx}
\usepackage{algpseudocode}
\usepackage{graphicx}

\usepackage{booktabs}
\usepackage{colortbl}
\usepackage{color}

\title{Beer Organoleptic Optimisation:\\ \large Utilising Swarm Intelligence and Evolutionary Computation Methods}

\author{
Mohammad Majid al-Rifaie\thanks{Corresponding author.} \hspace{0.3cm}
Marc Cavazza \vspace{5mm} 
\\ 
University of Greenwich\\
School of Computing and Mathematical Sciences
\\
\small{\texttt{ \{M.AlRifaie, M.Cavazza\} @ gre.ac.uk }}
}

\nipsfinalcopy 

\begin{document}
\maketitle

\begin{abstract}
Customisation in food properties is a challenging task involving optimisation of the production process with the demand to support computational creativity which is geared towards ensuring the presence of alternatives. This paper addresses the personalisation of beer properties in the specific case of craft beers where the production process is more flexible. We investigate the problem by using three swarm intelligence and evolutionary computation techniques that enable brewers to map physico-chemical properties to target organoleptic properties to design a specific brew. While there are several tools, using the original mathematical and chemistry formulas, or machine learning models that deal with the process of determining beer properties based on the pre-determined quantities of ingredients, the next step is to investigate an automated quantitative ingredient selection approach. The process is illustrated by a number of experiments designing craft beers where the results are investigated by ``cloning'' popular commercial brands based on their known properties. Algorithms performance is evaluated using accuracy, efficiency, reliability, population-diversity, iteration-based improvements and solution diversity. The proposed approach allows for the discovery of new recipes, personalisation and alternative high-fidelity reproduction of existing ones.

\end{abstract}

\section{Introduction}
\label{sec:introduction}
The optimisation of food production processes, besides its real-world significance, is faced with the apparently contradictory challenge of finding solutions to meeting precise characteristics as well as offering some diversity of solution which reconstruct the diversity of tastes and preferences.
Given the presence of several viable solutions when optimising food processes, this real-world problem poses itself as a challenging task with an inherently underdetermined characteristic~\cite{donoho2012sparse,phillips2014best}. In this work, we propose swarm intelligence and evolutionary computation techniques as the means to identify high quality and diverse solutions. This paper applies three population-based algorithms -- particle swarm optimisation~(PSO)~\cite{Kennedy_Eberhart_1995}, dispersive flies optimisation~(DFO)~\cite{alRifaie_2014_DFO}, and differential evolution~(DE)~\cite{Storn_Price_1995} -- for optimising beer recipes based on pre-determined organoleptic properties. The complexity of the brewing process necessitates an often strict adherence to existing recipes and the associated instructions with the aim of reducing mishap chances and to avoid costly guessworks~\cite{steenackers2015chemical}; this is especially the case when the primary goal is the production of a beer with particular organoleptic characteristics. 

This work enables the use of an automated \textit{quantitative ingredients selection}, which as of today, constitutes one of the primary experimental aspects of brewing.
In this paper, Section~\ref{sec:background} presents previous and related work, followed by introducing some key concepts, terminology and formulas which determine the fermentation process, from which the fitness value for the optimisation methods is determined. This is then followed by presenting the three swarm intelligence and evolutionary computation methods in Section~\ref{sec:methods}. Subsequently, Section~\ref{sec:experiments} proposes several experiments along with the experiment setup and performance measures to evaluate the performance of the optimisers with real-world input. Section~\ref{sec:results} reports on the experiments results and provides discussion on the algorithms' performance when optimising three ``cloned'' beer properties over the performance measures, solution vectors diversity, iteration-based improvements and solution clustering. Finally, the paper is concluded by presenting ongoing and future work.

\section{Background}
\label{sec:background}

The process of beer brewing has attracted various attempts at optimising or automating different elements of the process. These have however most often considered specific relationships or causal relationships between ingredients and isolated properties known to play a significant role in consumers' preferences (e.g. foamability). 
Ermi et al.~\cite{ermi2018deep} explore two deep learning architectures  to  model  the  non-linear  relationship  between  beer in  these  two  domains with the aim of  \textit{classifying}  coarse- and  fine-grained  beer type  and  \textit{predicting  ranges}  for  original  gravity,  final  gravity, alcohol by volume, international bitterness units and colour\footnote{Note that ABV is a function of OG and FG (see Section~\ref{subsec:formulas}).}. 

Another research is conducted for beer foamability~\cite{viejo2018robotics} where robotics and computer vision techniques are combined with non-invasive consumer biometrics to assess quality traits from beer foamability. Furthermore, in another study~\cite{gonzalez2018assessment}, an objective predictive model is developed to investigate the intensity levels of sensory descriptors in beer using the physical measurements of colour and foam-related parameters where a robotic pourer, was used to obtain some colour and foam-related parameters from a number of different commercial beer samples. It is claimed that this method could be useful as a rapid screening procedure to evaluate beer quality at the end of the production line for industry applications.

Using various AI techniques, several other predictive studies are presented concerning fermentation, monitoring and control~\cite{mileva2008ann,vassileva2010ai}, controlling of beer fermentation process using population-based optimisers~\cite{andres2004multiobjective}, 
predicting beer flavours~\cite{wilson2003application}, measurement and information processing in a brewery~\cite{campbell2003soft} and predicting aceticacid content in the final beer~\cite{zhang2012predicting}.

This work aims at utilising population-based methods in a way that would facilitate the discovery of variants or novel recipes for some target properties of the brew. 

\begin{figure}
	\centering
	\fbox{\includegraphics[width=0.7\linewidth]{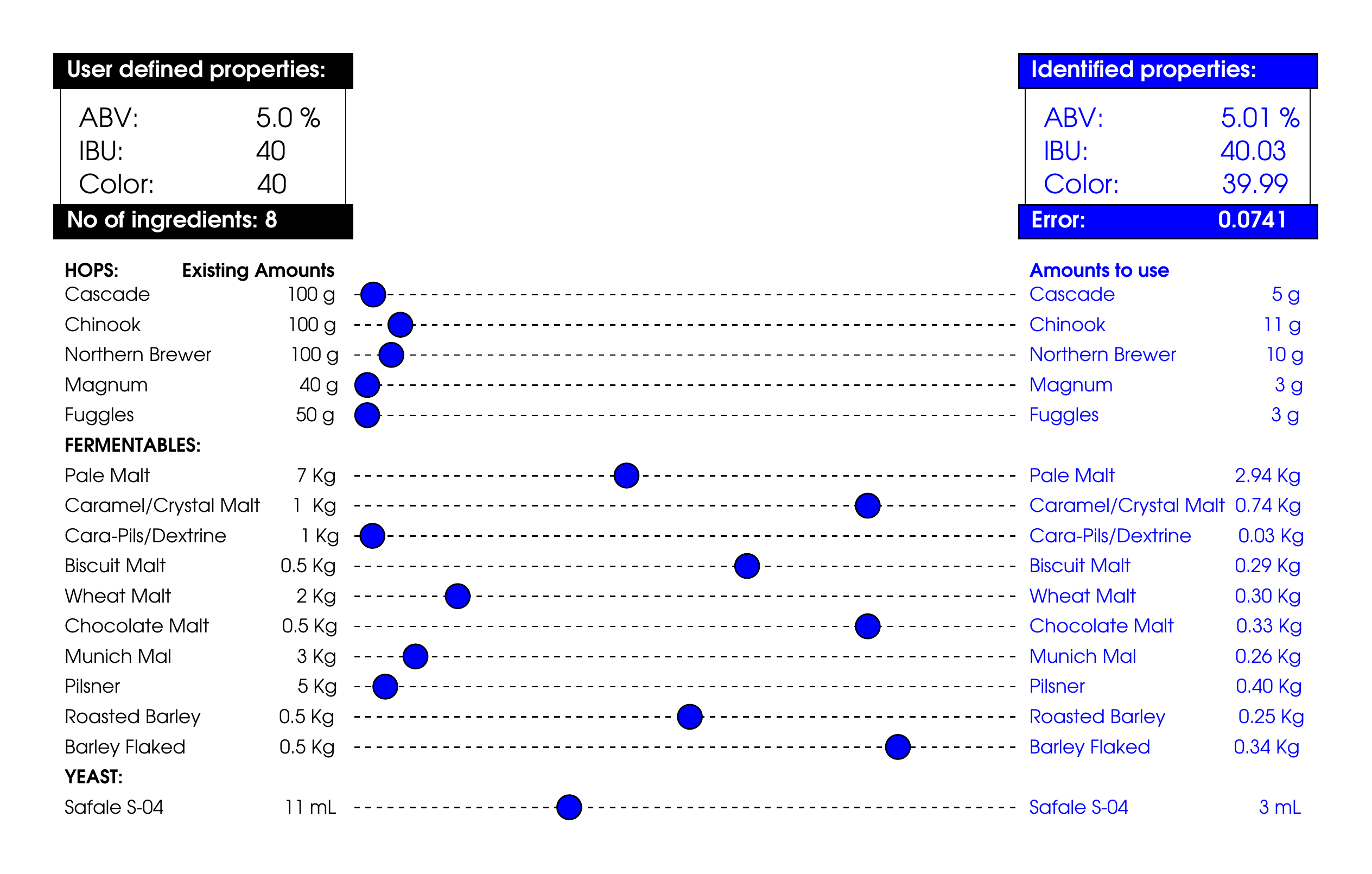}}
	\caption{Schematic view of the brewing process optimisation. The control panel on top-left corner takes users' desired values, and the top-right panel shows the corresponding optimal values found so far based on the ingredients in the inventory. The lines represent each of the in-stock items and the circles indicate the suggested quantities. }
	\label{fig:AI_Beer}
\end{figure}

\subsection{Process equations \& optimisation variables}\label{subsec:formulas}

In the brewing process, ingredients are divided in three broad categories: hops, fermentables or yeasts. In addition to weight, several other relevant features are also needed to calculate their impact in the brewing process (e.g. hop's alpha and beta; fermentable's yield, colour, moisture and diastatic power; yeast's minimum and maximum temperatures, and attenuation). Beer's taste changes significantly depending on the exact quantities and varieties of ingredients and their timing in the process. 
The key physio-chemical properties which contribute towards computing the fitness value of the solutions are alcohol by volume (ABV), bitterness (IBU) and colour which are used by the optimiser to determine the suitability of each proposed solution.
From a food science perspective, the brewing process, although in some parts empirical, has been the subject of many descriptions and partial formalisation which are however sufficient to derive relevant equations. More specifically, a number of formal relationships between ingredients and target organoleptic properties are sufficiently specific to support the generation of fitness functions. Some of the relevant formulas are discussed next.

\textbf{ABV} $=f(\text{OG},\text{FG})$ and is defined as~\cite{papazian1991new}:
\begin{eqnarray}
\text{ABV} = 131.25 \times (\text{OG} - \text{FG})
\end{eqnarray}

When ABV is above $6$ or $7\%$ the following is used which provides a higher level of accuracy~\cite{hall1995brew,daniels1998designing}: 
\begin{eqnarray}
\text{ABV} =\frac{76.08~(\text{OG}-\text{FG})\text{FG}} {0.794~(1.775-\text{OG})}
\end{eqnarray}

\textbf{IBU}
is determined by taking into account the bitterness produced by hops or the hop extracts (from the fermentables), thus $\text{IBU} = f(\vec{\text{hops}},\vec{\text{fermentables}},\text{volume})$. The bitterness produced by hop is calculated as follows: 
\begin{eqnarray}
\text{IBU}_h = \sum_{i = 1}^{N_h} \frac{10w_i\alpha_i(1 - \exp^{-0.04 t_i})}{4.15~v} 1.65 \times 0.000125^{(\text{OG}-1)} 
\end{eqnarray}
where $N_h$ is the number of hops; $w$ represents the weight; $v$ is the volume or batch size; $t$ is time in minutes; and fermentables' bitterness is defined as:
\begin{eqnarray}
\text{IBU}_{f} = \sum_{i = 1}^{N_f} \frac{g_i~w_i}{v}
\end{eqnarray}
\noindent where $N_f$ is the number of fermentables; and $g$ is `IBU gal per lb' which is associated with each fermentable and is known for each ingredients.
The final IBU is the sum of the individual IBUs: $\text{IBU} = \text{IBU}_h + \text{IBU}_f$.

\noindent\textbf{IBU/GU}
is often described in the following categories: cloying, slightly malty, balanced, slightly hoppy, extra hoppy, and very hoppy. $\text{IBU/GU} = f(\text{OG},\text{IBU})$:
\begin{eqnarray}
\text{IBU/GU} = \frac{\text{IBU}}{1000(\text{OG} - 1)} 
\end{eqnarray}
\textbf{Colour} is mainly determined by malts and hops.
The two main protocols to measure colour are Standard Reference Method~(SRM) and European Brewing Convention~(EBC). 
Table~\ref{tab:colours} shows representative colours. 
SRM, which is used in this work, was initially adopted in 1950 by the American Society of Brewing Chemists. 
The value of SRM is determined by measuring the attenuation of light of a particular wavelength ($430$ nm) in passing through $1$ cm of the beer, expressing the attenuation as an absorption and scaling the absorption by a constant ($12.7$ for SRM or $25$ for EBC, where $\text{EBC} = \text{SRM} \times 1.97$).

Stone and Miller~\cite{stone1949standardization} proposed malt colour unit~(MCU), which is defined as:
\begin{eqnarray} \label{eq:MCU}
	\text{MCU} = \sum_{i=1}^{N_f} \frac{c_i~ w_i}{v}
\end{eqnarray}
\noindent where $c$ refers to grains' colour (fermentables' colour). 
As shown in the equation above, for more than one grain type, the MUC is calculated for each and all the values are summed.

However, MUC tends to overestimate the colour value for darker beers (MUC $> 10.5$). 
Thus, Morey~\cite{morey2000hop} derived an equation to deal with SRM up to 50: 
\begin{eqnarray}\label{eq:COLOUR}
\text{SRM} = 1.4922 \sum_{i=1}^{N_f} \frac{c_i~ w_i^{0.6859}}{v}
\end{eqnarray}
\noindent where $c$ refers to grains' colour (fermentables' colour).

\begin{table} 
	\centering \footnotesize
	\setlength{\tabcolsep}{7pt}
	\begin{tabular}{lccl}  
		\toprule
		& \textbf{SRM} & \textbf{EBC} & \textbf{Colour} \\
		\midrule
		{\cellcolor[HTML]{FFFF45}}  & 2 & 4 & Pale Straw      \\
		{\cellcolor[HTML]{FFE93E}}  & 3 & 6 & Straw      \\
		{\cellcolor[HTML]{FEd849}}  & 4 & 8 & Pale Gold      \\
		{\cellcolor[HTML]{FFA846}}  & 6 & 12 & Deep Gold      \\
		{\cellcolor[HTML]{F49F44 }}  & 9 & 18 & Pale Amber      \\
		{\cellcolor[HTML]{D77F59 }}  & 12 & 24 & Medium Amber      \\
		{\cellcolor[HTML]{94523A }}  & 15 & 30 & Deep Amber      \\
		{\cellcolor[HTML]{804541 }}  & 18 & 35 & Amber-Brown      \\
		{\cellcolor[HTML]{5b342F }}  & 20 & 39 & Brown      \\
		{\cellcolor[HTML]{4C3b2B }}  & 24 & 47 & Ruby Brown      \\
		{\cellcolor[HTML]{38302E }}  & 30 & 59 & Deep Brown      \\
		{\cellcolor[HTML]{31302C }}  & 40+ & 79 & Black      \\
		\bottomrule
	\end{tabular}
	\caption{Beer colour in SRM and EBC values}
	\label{tab:colours}
\end{table}

One of the key contributions of this work is the application of a suite of population-based algorithms which take an in-stock inventory of existing ingredients and their quantities (see Table~\ref{tab:stockII}) along with a desired set of physico-chemical features of a beer, and as output return an optimal set of ingredient list and their associated quantities which facilitate the production of a target beer with the desired organoleptic properties (see Figure~\ref{fig:AI_Beer}).

\section{Swarm and evolutionary methods}
\label{sec:methods}
The algorithms used in this work are 
particle swarm optimisation (PSO)~\cite{Kennedy_Eberhart_1995} as one of the most well-known swarm intelligence algorithms; Differential evolution (DE)~\cite{Storn_Price_1995}, a well-known and efficient evolutionary computation method; and a minimalist component-stripped swarm optimiser, dispersive flies optimisation (DFO)~\cite{alRifaie_2014_DFO}, which solely relies on particles' positions at time $t$ to generate the positions for time  $t+1$ (therefore not using additional vectors, such as PSO's memory and velocity, or DE's mutant and trial vectors)\footnote{
	It was demonstrated that despite DFO's simplicity, it exhibits a competitive performance when compared with the standard versions of PSO~\cite{Kennedy_Eberhart_1995},
	GA~\cite{Goldberg1989} and DE~\cite{Storn_Price_1995} on a set of $28$
	benchmarks over three performance measures of error, efficiency and
	reliability~\cite{alRifaie_2014_DFO}. It was shown that DFO is more efficient in 85\% and more reliable in 90\% of the $28$ standard optimisation benchmarks used;
	furthermore, when there exists a statistically significant difference,
	DFO converges to better solutions in 71\% of problem set. Furthermore, DFO has been applied to various problems, including but not limited to medical imaging~\cite{alRifaie_2016}, optimising machine learning algorithms~\cite{Haya_Thesis_2018,Alhakbani_2017_DFO_SVM}, training deep neural networks for false alarm detection in intensive care units~\cite{Hooman_2018_Deep_DFO}, computer vision and quantifying symmetrical complexities~\cite{alRifaie_2017_EvoMus_DFO_SymComplexity}, identifying  animation key points from medialness maps~\cite{Aparajeya_EvoMusART_2019} and analysis of autopoiesis in computational creativity~\cite{alRifaie_2017_autopoiesis_DFO}.
	
	DFO's source code can be found at \scriptsize \url{https://github.com/mohmaj/DFO}
	}. 
The standard versions of these algorithms are used, therefore allowing performance comparisons between these simple yet powerful optimisers. 
For each of these algorithms the position vector of each member of the population is defined as:
\begin{equation}
\vec{x}_{i}^{t} = \left[x_{i0}^{t}, x_{i1}^{t}, ..., x_{i,D-1}^{t}\right],\qquad 
i \in \{0,1,2,...,\textsl{N-1}\}
\end{equation}
\noindent where $i$ represents the $i^\text{th}$ individual, $t$ is
the current time step, $D$ is the problem dimensionality,
and $N$ is the population size. For continuous problems, $x_{id} \in
\mathbb{R}$ (or a subset of $\mathbb{R}$). 

In the first iteration, where $t=0$, the $i^\text{th}$ member's $d^\text{th}$ component is initialised as:
\begin{equation}
\label{eq:init}
x_{id}^{0} = \text{U}(x_{\text{min},d}, x_{\text{max},d})
\end{equation}

The update equation of a standard version of each of the algorithms are provided below (PSO's Clerc-Kennedy, standard DFO, and DE/best/1):
\begin{eqnarray*} 
	\text{PSO}&:&v_{id}^{t+1}=\chi\left(v_{id}^{t}+c_{1}r_{1}\left(p_{id}-x_{id}^{t}\right)+c_{2}r_{2}\left(g_{id}-x_{id}^{t}\right)\right) \label{eq:PSO}\\
	&:& x_{id}^{t+1}=v_{id}^{t+1}+x_{id}^{t}\\
	&:& x_{id}^{t+1} = f(v_{id}^{t+1},\; p_{id},\; g_{d},\; x_{id}^{t})\label{eq:PSO_f}\\
	\nonumber \\
	\text{DFO}&:& x_{id}^{t+1} =  x_{i_nd}^{t} + u ( x_{sd}^{t} - x_{id}^{t} )\label{eq:DFO}\\
	&:& x_{id}^{t+1} = f(\vec{x}_{d}^{t}) \label{eq:DFO_f}\\
	\nonumber \\
	\text{DE}&:& v_{id}^{t+1} = x_{best,d}^{t} + F\left(x_{r_{1}d}^{t} - x_{r_{2}d}^{t}\right)\label{eq:DE2}\\
	&:& u_{id}^{t+1} = \left\{ \begin{array}{ll}
		v_{id}^{t}, \; & \mathrm{if} \;\;r \leq CR \;\; \mathrm{or} \; d = U(0,1)\\
		\\
		x_{id}^{t}, \; & \mathrm{otherwise}
	\end{array}\right.\\
	&:& x_{id}^{t+1} = f(v_{id}^{t+1},\; u_{id}^{t+1},\; \vec{x}_{d}^{t})\label{eq:DE2_f}
\end{eqnarray*}

\noindent where for PSO, $\chi$ is the constriction factor which is set to $0.72984$~\cite{Bratton_Kennedy_2007}; $v_{id}^{t}$
is the velocity of particle $i$ in dimension~$d$ at time
step $t$; $c_{1,2}$ are the learning factors (also referred to
as acceleration constants) for personal best and neighbourhood best
respectively; ${r}_{1,2}$ are random numbers adding stochasticity to the algorithm
and they are drawn from a uniform distribution on the unit interval
$U\left(0,1\right)$; ${p}_{id}$ is the personal best position
of particle $\vec{x}_{i}$ in dimension $d$; and $g_{d}$ is swarm best at dimension $d$. In DFO, which uses a ring topology, $\vec{x}_{i_n}$ is the position of $\vec{x}_i$'s best \textit{neighbouring} individual, $\vec{x}_{s}$ is the position of \textit{swarm}'s best individual where $s \in \{0,1,2,...,\textsl{N-1}\}$, ${u}$ is a random number drawn from a uniform distribution on the unit interval $U\left(0,1\right)$; the diversity of population in DFO is maintained by a component-wise jump which is triggered when $U(0,1)<\Delta$ where $\Delta=0.001$. 
For DE's mutant vector (DE/best/1), $v_{id}$ is $d^{\text{th}}$ gene of the $i^\text{th}$ chromosome's mutant vector\footnote{Vector $\vec{v}$ in PSO and DE are different, albeit they carry the same name in the literature.}; $u_{id}$ is $d^{\text{th}}$ gene of the $i^\text{th}$ chromosome's trial vector; $r_{1}$ and $r_{2}$ are different from $i$ and are distinct random integers drawn from the range $\left[0,N-1\right]$; and $x_{best,d}^{t}$ is the $d^\text{th}$ gene of the best chromosome at generation $t$; and $F$ is a positive control parameter for constricting the difference vectors which is set to $0.5$. The crossover operation in DE, improves population diversity through exchanging some components using the crossover rate (CR), which is set to $0.9$. Elitism is used for DFO and DE, with an elite size of one maintaining the best found solution. In this work, if the updated position for a dimension is outside the boundaries, its value is clamped to the edges.

\section{Experiments}
\label{sec:experiments}
This section presents a set of experiments where physico-chemical properties of three commercial beers (i.e. Guinness Extra Stout, Kozel Black, Imperial Black IPA) are used along with the in-stock inventory to evaluate the proposed system by ``reverse manufacturing'' these commercial beers from their target physico-chemical properties.
Figure~\ref{fig:AI_Beer} shows the schematic view of the developed system with regard to user input vectors and expected vector output.
The list of ingredients in this experiment is shown in Table~\ref{tab:stockII}, and the desired physio-chemical properties for this set of experiments are derived from three existing commercial beers as shown in Table~\ref{tab:RealBeerFeatures}.

The experiments reported in this section, compare the results of the optimisers over each product. This is then followed by another set of experiments investigating the behaviour of the algorithms in terms of iteration-based improvements throughout the optimisation process. The solution vectors diversity for each of the optimisers over each product is investigated. Additionally, to further evaluate the solution vectors diversity, distinct solution clusters are generated by each algorithm and for each product.

\begin{table}[t]
	\centering \footnotesize 
	\begin{tabular}{lllr}  
		\toprule
		& \textbf{Type} & \textbf{Name} & \textbf{Amount}\\
		\midrule
		1 & Hop & Cascade & 100 g\\
		2 &  & Chinook & 100 g\\
		3 &  & Northern Brewer & 100 g\\
		4 &  & Magnum & 40 g \\
		5 &  & Fuggles & 50 g\\
		\midrule
		6 & Fermentable & Pale Malt (UK) & 7 kg\\
		7 &  & Caramel/Crystal Malt  & 1 kg\\
		8 &  & Cara-Pils/Dextrine & 1 kg\\
		9 &  & Biscuit Malt & 0.5 kg\\
		10 &  & Wheat Malt  (Belgium) & 2 kg\\
		11 &  & Chocolate Malt (UK) & 0.5 kg\\
		12 &  & Munich Malt & 3 kg\\
		13 &  & Pilsner (German) & 5 kg\\
		14 &  & Roasted Barley & 0.5 kg\\
		15 &  & Barley  Flaked & 0.5 kg\\
		\midrule
		16 & Yeast & Safale S-04 & 11 mL\\
		\bottomrule
	\end{tabular}
	\caption{List in-stock inventory\label{tab:stockII}}
\end{table}

\begin{table}
	\centering \footnotesize 
	\begin{tabular*}{\columnwidth}{@{\extracolsep{\fill}} lrlll}
		\toprule
		\textbf{Name} & \textbf{ABV} & \textbf{IBU} & \textbf{SRM} & \textbf{Origin} \\
		\midrule
		Guinness Extra Stout & 5.00 \% & 40 & 40 & Dublin, Ireland\\
		Kozel Black & 3.80 \% & 15 & 24 & Prague, Czech\\
		Imperial Black IPA & 11.20 \% & 150 & 35 & Ellon, Scotland\\
		\bottomrule
	\end{tabular*}
	\caption{Sample beer characteristics in three products\label{tab:RealBeerFeatures}}
\end{table}

\subsection{Experiment setup}

In order to set up the simulation experiment, a realistic inventory of a home brewer in Table~\ref{tab:stockII} along with physio-chemical properties of three existing commercial beers in Table~\ref{tab:RealBeerFeatures} are used as the benchmark and the analyses are investigated on that basis. 
In these experiments, the population size for each algorithm is set to $100$ and termination criterion is set to either reaching $150,000$ function evaluations (FEs) or reaching a corresponding error depending on the product being optimised, with error less than or equal to $0.05899,~0.070560,~0.00498$, for Guinness Extra Stout, Kozel Black and Imperial Black IPA respectively\footnote{These values are the best found values irrespective of the algorithm choice or number of function evaluations and are therefore used as the base optima.}. 
There are 50 Monte Carlo simulations for each experiment and the results are summarised
over these independent simulations. 

\subsection{Performance measures\label{subsec:measures}}
In order to measure the presence of any statistically significant differences in the performance of the algorithms, and for pairwise statistical comparisons, Wilcoxon $1\times1$ non-parametric statistical test is deployed~\cite{wilcoxon1970critical}. The performance measures used in this paper are error, efficiency, reliability and diversity. 
\emph{Error} or accuracy is defined by the quality of the best member in terms of its closeness to the optimum position (i.e. minimisation). 
\begin{eqnarray}
\textsc{Error}=f(\vec{x}) = \sum_{i=1}^{N_p} \sqrt{(f_{p_i}(\vec{x}) - p^*_i)^2 }
\end{eqnarray}
\noindent where $\vec{x}$ is the list of ingredients and $N_p = 3$ is the number of parameters; $p_1$: ABV, $p_2$: IBU, $p_3$: Colour, where the relevant equations are provided in Section~\ref{subsec:formulas}; $p^*_i$ represents the desired value provided by the brewers and the termination criterion for each run is dependent on the problem itself.  Another measure used is \emph{efficiency} which is the number of function evaluations before reaching a specified error, and \emph{reliability} is the percentage of trials where a specified error is reached. 
\begin{eqnarray}
\textsc{Efficiency}&=&\frac{1}{n}\sum_{i=1}^{n}\textrm{FEs},\label{eq:Swarm Efficiency}\\
\textsc{Reliability}&=&\frac{n^{'}}{n}\times100 \label{eq:Swarm Reliability}
\end{eqnarray}
where \emph{n} is the number of trials in the experiment and $n^{'}$ is the number of successful trials.
Additionally, \textit{diversity}, is used to study the population's behaviour with regard to exploration and exploitation. There are various approaches to measure diversity. The average distance around the population centre is shown to be a robust measure in the presence of outliers~\cite{Olorunda_2008}:
\begin{eqnarray}
\textsc{Diversity}&=&\frac{1}{N}\sum_{i=1}^{N}\sqrt{\sum_{d=1}^{D}\left(x_{id} - \bar{x}_d\right)^2},\label{eq:Swarm Diversity}\\
\bar{x}_d&=&\frac{1}{N}\sum_{i=1}^{N}x_{id}
\end{eqnarray}
\noindent where $N$ is the population size, and $\bar{x}_d$ is the average value of dimension $d$ over all members of the population.
For these experiments, the brewer's efficiency is set to $70\%$\footnote{Efficiency, in home-brewing context, indicates how efficient the equipment and processes are in extracting sugars from the malts during the mash stage.}, boil size of $24$L, batch size of $20$L and boil time is set to $60$ minutes.

\begin{figure}
	\newlength{\imgW}
	\setlength{\imgW}{0.32\linewidth}
	
	\centering
	Guinness Extra Stout \hspace{0.15\textwidth} Kozel Black \hspace{0.15\textwidth} Imperial Black IPA
	\hspace{0.04\textwidth}
	
	\includegraphics[width=\imgW]{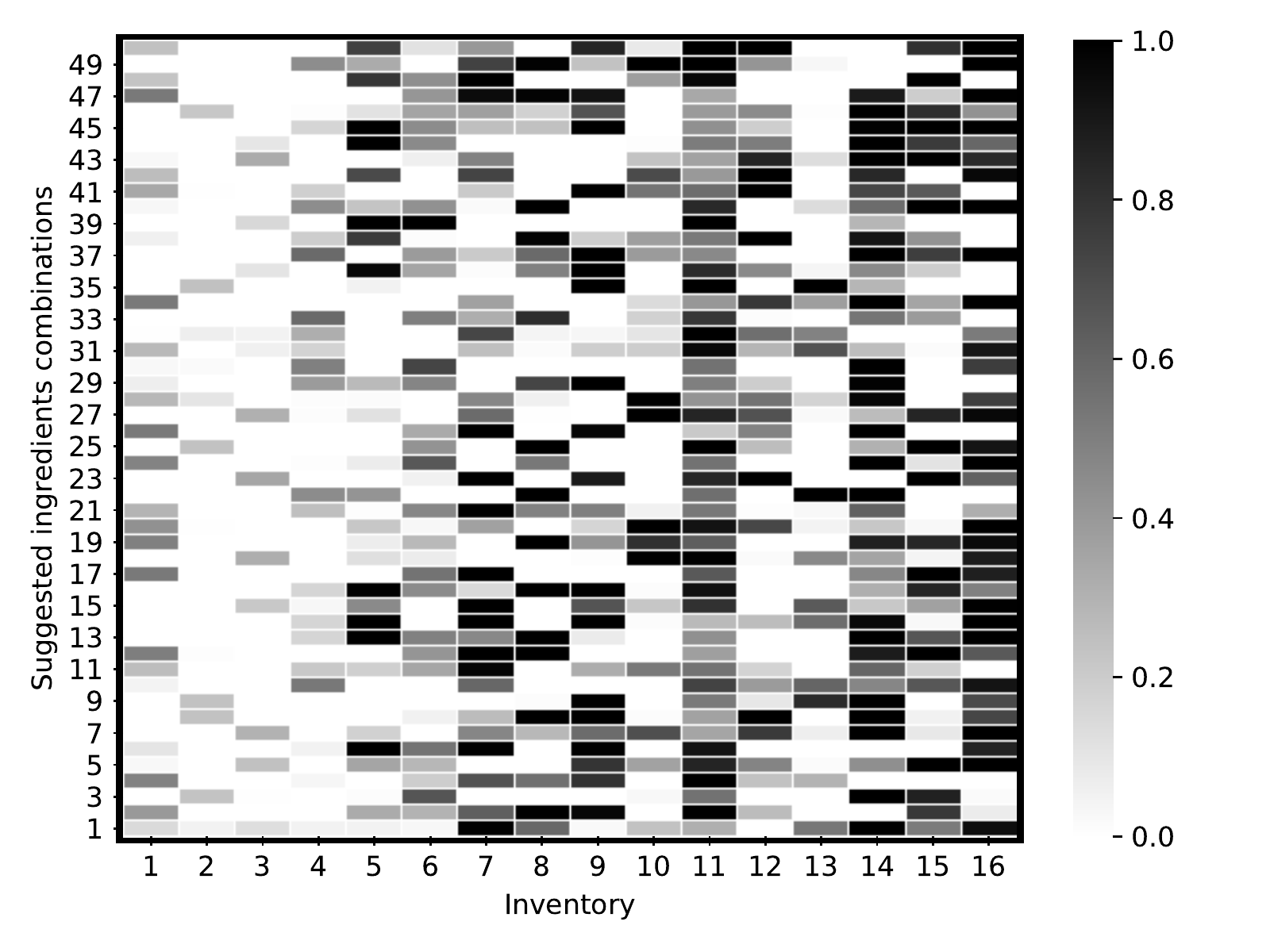} 
	\includegraphics[width=\imgW]{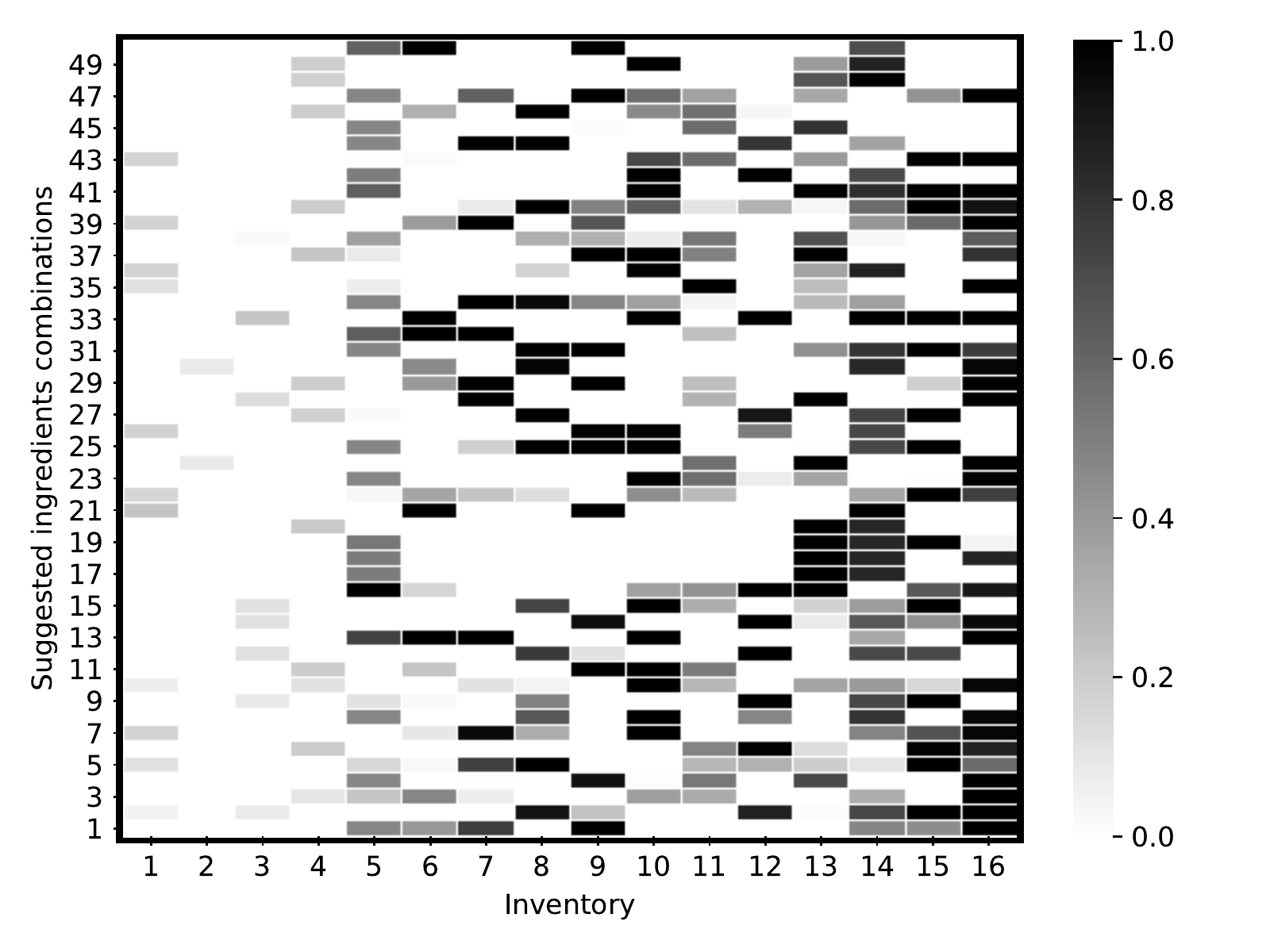} 
	\includegraphics[width=\imgW]{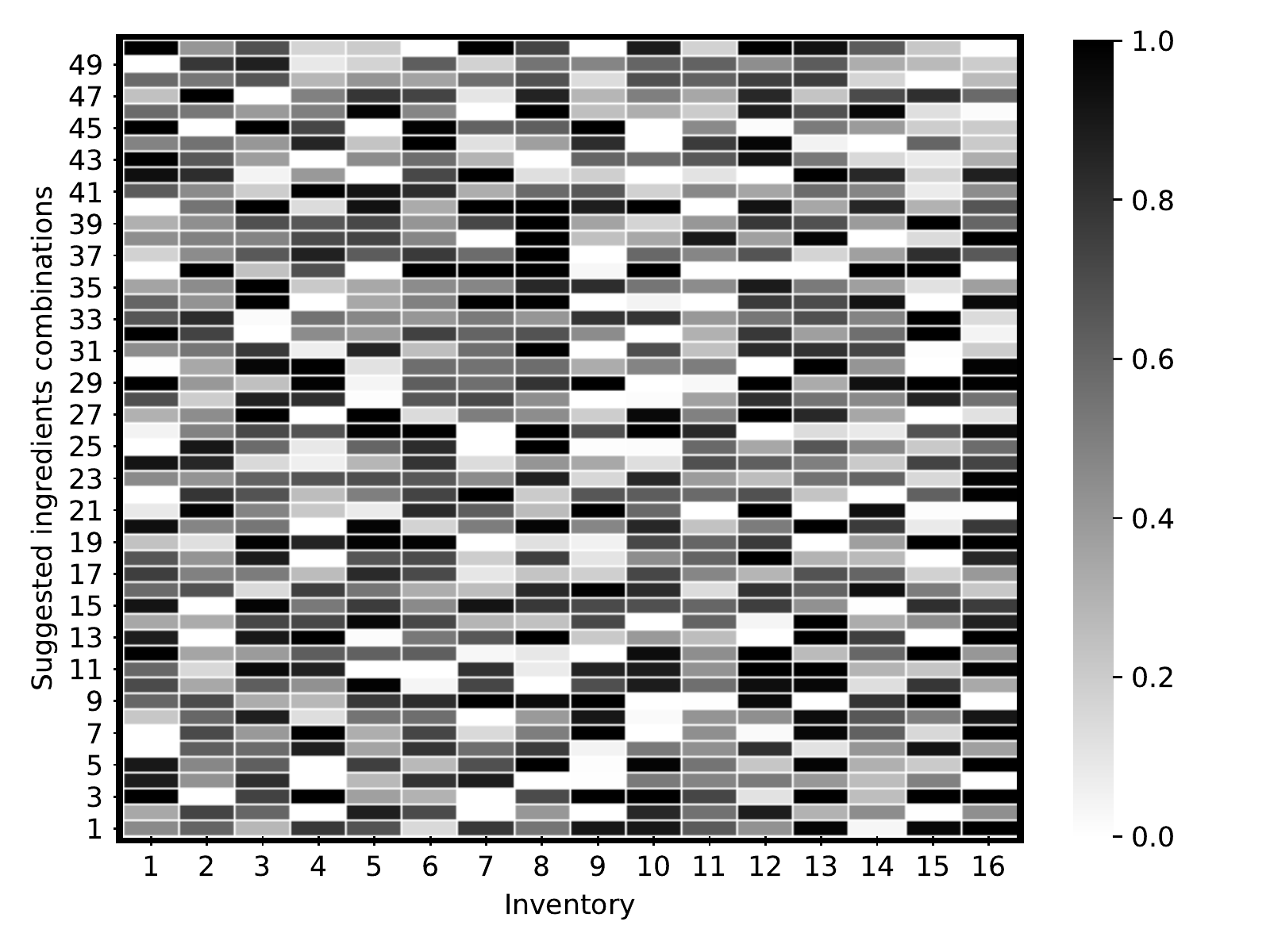}

	\caption{Ingredients combinations generated by PSO for three products, illustrating recommended ingredients uptake proportion, as well as independent solutions' diversity for each of the product.\label{fig:PSO_sols}}
\end{figure}

\section{Results and discussion\label{sec:results}}
This section reports the results outlined in the experiments section where algorithms' performances are contrasted using the performance measures, as well as iteration-based improvements. This is then followed by investigating the diversity of the solution vectors which are generated by each optimiser for each product, as well as studying the distinct solution clusters within each optimiser-product pair.
To demonstrate the process, Figure~\ref{fig:PSO_sols} illustrates $50$ solution vectors for each of the products (i.e. Guinness Extra Stout, Kozel Black and Imperial Black IPA) which are generated by PSO.
These vectors visualise various viable ingredients combinations and the uptake of each of the $16$ ingredients when reaching the termination point.

\subsection{Accuracy, efficiency, reliability \& diversity\label{subsec:acc}}
Algorithms performance are initially compared over each product independently; this is then followed by summarising the findings over all products.
When optimising Guinness Extra Stout, in the~50~independent trials (Table~\ref{tab:G-err}) all three algorithms, in some or all trials, reach the optimum error. 
In cases where the optimum is not found, PSO returns the highest error followed by DE (see Error$\rightarrow$Worst in Table~\ref{tab:G-err}). 
In terms of efficiency (Efficiency$\rightarrow$Mean), PSO is shown to be requiring the largest number of function evaluations in the given problem, followed by DE. In other words, DFO is around twice as efficient as DE, which in turn is approximately twice as efficient as PSO. 

\begin{table}[t]	
	\centering \footnotesize
	\begin{tabular}{lllll}
		\toprule 
		&  & {PSO} & {DFO} & {DE}\tabularnewline
		\midrule 
		{Error} & Best & 0.0590 & 0.0590 & 0.0590\tabularnewline
		& Worst & 1.7852 & 0.0870 & 0.1062\tabularnewline
		& Median & 0.0590 & 0.0590 & 0.0590\tabularnewline
		& Mean & 0.1209 & 0.0595 & 0.0599\tabularnewline
		& StDev & 0.2795 & 0.0040 & 0.0067\tabularnewline
		\midrule 
		{Efficiency} & Best & 9900 & 3000 & 6766\tabularnewline
		& Worst & 24800 & 17200 & 11940\tabularnewline
		& Median & 17300 & 3800 & 8557\tabularnewline
		& Mean & 17427.66 & 4389.80 & 8678.84\tabularnewline
		& StDev & 3498.30 & 2460.46 & 1184.86\tabularnewline
		\midrule 
		{Diversity} & Successful & 0.9280 & 0.8125 & 0.0745\tabularnewline
		& Failed & 4.75E-04 & 2.24E-05 & 3.36E-15\tabularnewline
		\midrule 
		{Reliability} & Reliability  & 47 (94\%) & 49 (98\%) & 49 (98\%)\tabularnewline
		\bottomrule 
	\end{tabular}
	\caption{Guinness Extra Stout: algorithms performance for reverse brewing of the commercial beer\label{tab:G-err}}
\end{table}

\begin{table}
	\centering \footnotesize
	\begin{tabular}{lllll}
		\toprule 
		&  & {PSO} & {DFO} & {DE}\tabularnewline
		\midrule  
		{Error} & Best & 0.0706 & 0.0706 & 0.0706\tabularnewline
		& Worst & 11.4080 & 0.0706 & 8.6517\tabularnewline
		& Median & 0.0706 & 0.0706 & 0.0706\tabularnewline
		& Mean & 1.3400 & 0.0706 & 0.5560\tabularnewline
		& StDev & 2.5999 & 0.0000 & 1.5884\tabularnewline
		\midrule 
		{Efficiency} & Best & 8200 & 2900 & 6368\tabularnewline
		& Worst & 19400 & 13300 & 11741\tabularnewline
		& Median & 13000 & 4000 & 7960\tabularnewline
		& Mean & 13357.58 & 4482.00 & 8348.52\tabularnewline
		& StDev & 2810.03 & 1915.00 & 1202.83\tabularnewline
		\midrule 
		{Diversity} & Successful & 0.3375 & 0.6656 & 0.0183\tabularnewline
		& Failed & 6.72E-04 & -- & 3.15E-14\tabularnewline
		\midrule 
		{Reliability} & Reliability  & 34 (68\%) & 50 (100\%) & 42 (84\%)\tabularnewline
		\bottomrule 
	\end{tabular}
	\caption{Kozel Black: algorithms performance for reverse brewing of the commercial beer\label{tab:K-err}}
\end{table}

In terms of Kozel Black, as shown in Table~\ref{tab:K-err}, the algorithms exhibit the most varied performance in terms of error and reliability. 
While DFO reaches the optimum in all trials, PSO returns the highest error among the algorithms and shows the least reliability of 68\%. DFO exhibits efficiency outperformance, followed by DE. Considering the successful trials, DFO shows the highest diversity, followed by PSO while DE exhibits the least diversity (irrespective of whether the optimum is reached). 

The algorithms performance in terms of accuracy and reliability is comparable when optimising  Imperial Black IPA (see Table~\ref{tab:I-err}). In terms of the efficiency, the same trend continues, with DFO more than twice as efficient as DE, which is three times more efficient than PSO. Furthermore, PSO exhibits the largest FEs differences between successful trials. A potential contributing factor could be PSO's highest population diversity, which is a subject of an ongoing research.

\begin{table}	
	\centering \footnotesize
	\begin{tabular}{lllll}
		\toprule 
		&  & {PSO} & {DFO} & {DE}\tabularnewline
		\midrule 
		{Error} & Best & 0.0050 & 0.0050 & 0.0050\tabularnewline
		& Worst & 0.0050 & 0.0050 & 0.0050\tabularnewline
		& Median & 0.0050 & 0.0050 & 0.0050\tabularnewline
		& Mean & 0.0050 & 0.0050 & 0.0050\tabularnewline
		& StDev & 0.0000 & 0.0000 & 0.0000\tabularnewline
		\midrule 
		{Efficiency} & Best & 28100 & 4900 & 12139\tabularnewline
		& Worst & 73000 & 11800 & 19701\tabularnewline
		& Median & 49850 & 6250 & 16915\tabularnewline
		& Mean & 51022.00 & 6410.00 & 16739.88\tabularnewline
		& StDev & 10347.91 & 1110.48 & 1442.11\tabularnewline
		\midrule 
		{Diversity} & Successful & 1.3158 & 0.8767 & 0.0515\tabularnewline
		& Failed & -- & -- & --\tabularnewline
		\midrule 
		{Reliability} & Reliability  & 50 (100\%) & 50 (100\%) & 50 (100\%)\tabularnewline
		\bottomrule 
	\end{tabular}
	\caption{Imperial Black IPA: algorithms performance for reverse brewing of the commercial beer\label{tab:I-err}}
\end{table}

While these observations are representative of the algorithms performance, it is also important to identify areas with statistically significant differences between the algorithms. 
Using Wilcoxon test, Table~\ref{tab:summary-err-eff-rel}b demonstrates that DFO is the most efficient algorithm with a statistically significant difference from the other algorithms, and DE in the second place. This finding confirms the efficiency-related results reported in Tables~\ref{tab:G-err}, \ref{tab:K-err} and \ref{tab:I-err}. In all instances, DFO is, at least, twice as efficient as
DE, which in turn is, at least, 1.5 times more efficient than PSO (in Black
Kozel, and 3 times more efficient in Imperial Black IPA). 

Although the same trend continues for accuracy and reliability (see Tables~\ref{tab:summary-err-eff-rel}a and \ref{tab:summary-err-eff-rel}c),
more similarities between the algorithms are observed; for instance,
there are no statistically significant differences between the accuracy outcomes
when optimising Guinness Extra Stout, or Imperial Black IPA. Furthermore,
in terms of reliability, the algorithms exhibit consistent behaviour
when optimising Imperial Black IPA, all reaching the optimum accuracy
in all trials. 

Speculating the reason behind the algorithms' different performances, further studying the population diversity over different products could be helpful. For instance, when optimising Kozel Black (Table~\ref{tab:K-err}), population diversity in PSO and DE shrink by nearly a factor of $3$ and $4$ respectively from the diversity of successful population in the first product (Table~\ref{tab:G-err}), or by a factor of $4$ and $3$ from the third product (Table~\ref{tab:I-err}), while DFO maintains its near consistent population diversity. 
To better understand the algorithms' underlying performance, the next section studies the iteration-based improvements in each algorithm-product pair.

\begin{table} [t]
	\centering \footnotesize
	\begin{tabular}{lccc}
		\toprule 
		\textbf{(a) Error} & {PSO -- DFO} & {PSO -- DE\;} & {DFO -- DE\;}\tabularnewline
		\hline 
		{Guinness Extra Stout} & -- & -- & --\tabularnewline
		{Kozel Black} & o -- X & o -- X & X -- o\tabularnewline
		{Imperial Black IPA} & -- & -- & --\tabularnewline
		\bottomrule 
	\end{tabular}
	
	\vspace{3mm} 
	\begin{tabular}{lccc}
		\toprule 
		\textbf{(b) Efficiency }& {PSO -- DFO} & {PSO -- DE\;} & {DFO -- DE\;}\tabularnewline
		\hline 
		{Guinness Extra Stout} & o -- X & o -- X  & X -- o\tabularnewline
		{Kozel Black} & o -- X & o -- X  & X -- o\tabularnewline
		{Imperial Black IPA} & o -- X & o -- X  & X -- o\tabularnewline
		\bottomrule 
	\end{tabular}
	
	\vspace{3mm} 
	
	\begin{tabular}{lccc}
		\toprule 
		\textbf{(c) Reliability} & {PSO -- DFO} & {PSO -- DE\;} & {DFO -- DE\;}\tabularnewline
		\hline 
		{Guinness Extra Stout} & 0 -- 1 & 0 -- 1 & 1 -- 1\tabularnewline
		{Kozel Black} & 0 -- 1 & 0 -- 1 & 1 -- 0\tabularnewline
		{Imperial Black IPA} & -- & -- & --\tabularnewline
		\bottomrule 
	\end{tabular}
	\caption{Summary and statistical analysis. Based on Wilcoxon 1$\times$1 Non-Parametric Statistical Test, if the {error} or {efficiency} differences between each pair of algorithms are significant at the 5\% level, the pairs are marked. X -- o stands for the statistically significant performance of the left algorithm in comparison to the algorithm on the right. In terms of the \textit{reliability} measure, 0 -- 1 indicates that the right algorithm is more reliable.\label{tab:summary-err-eff-rel}}
\end{table}

\subsection{Iteration-based improvements\label{subsec:itrBased_impr}}

\begin{figure}
	\setlength{ \imgW }{.33\linewidth}

	\hspace{8mm} Guinness Extra Stout \hspace{0.16\textwidth} Kozel Black \hspace{0.16\textwidth} Imperial Black IPA 	

	\includegraphics[width=\imgW]{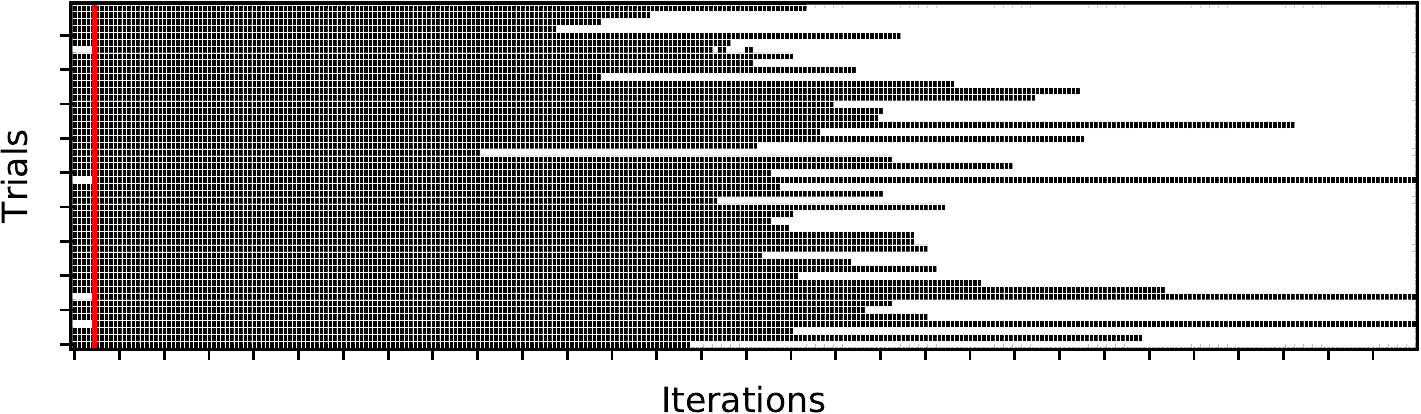} 
	\includegraphics[width=\imgW]{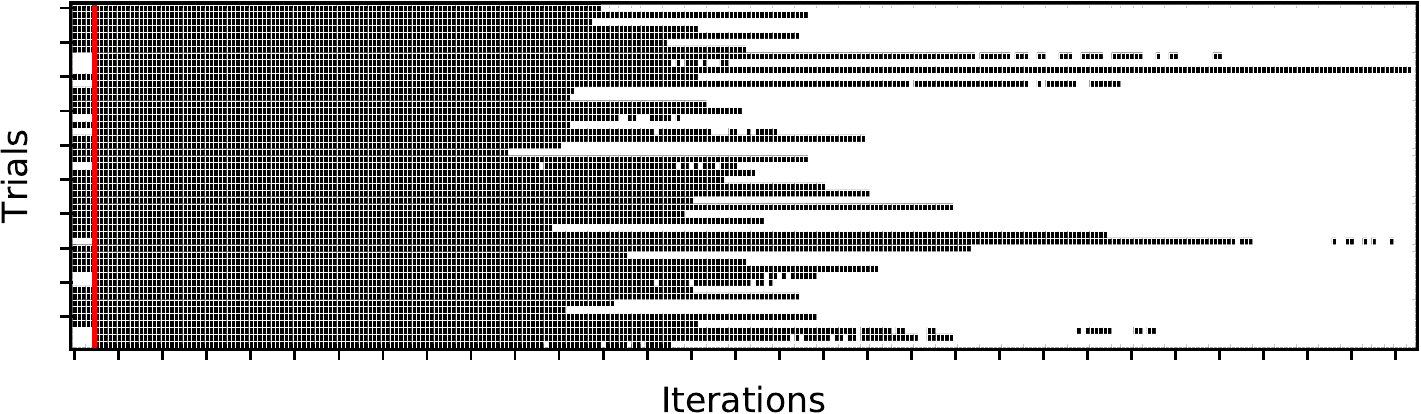}
	\includegraphics[width=\imgW]{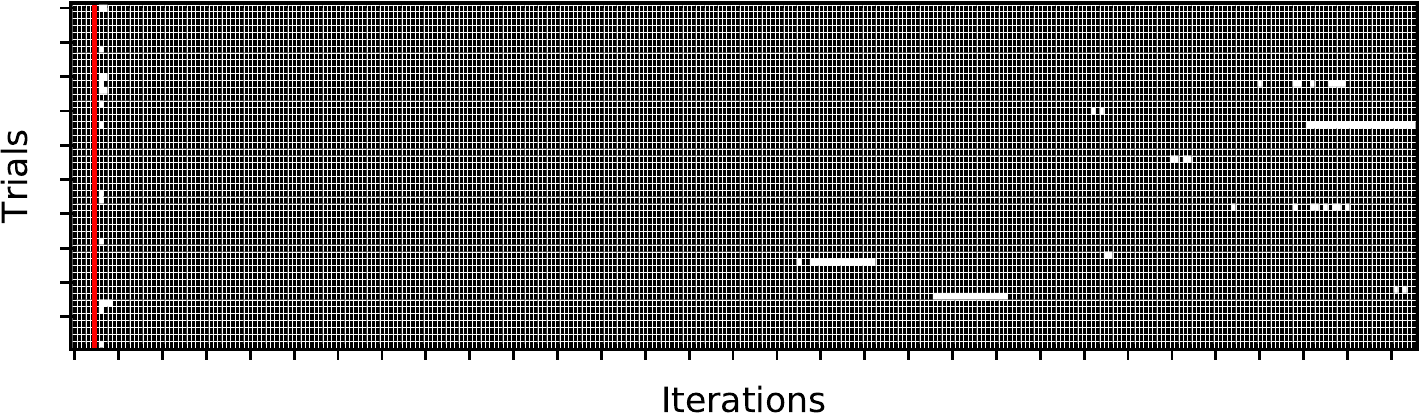}
	
	\includegraphics[width=\imgW]{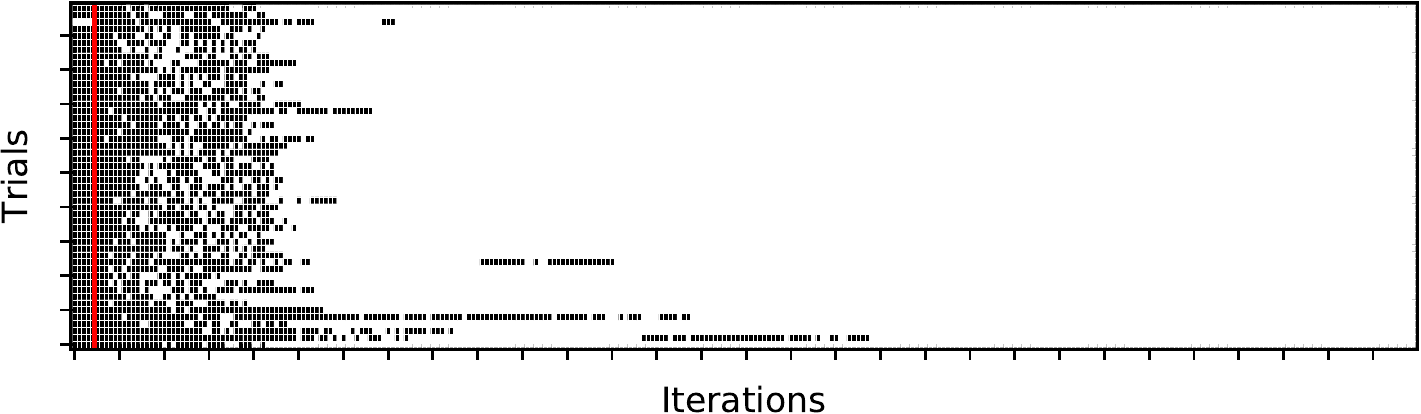} 
	\includegraphics[width=\imgW]{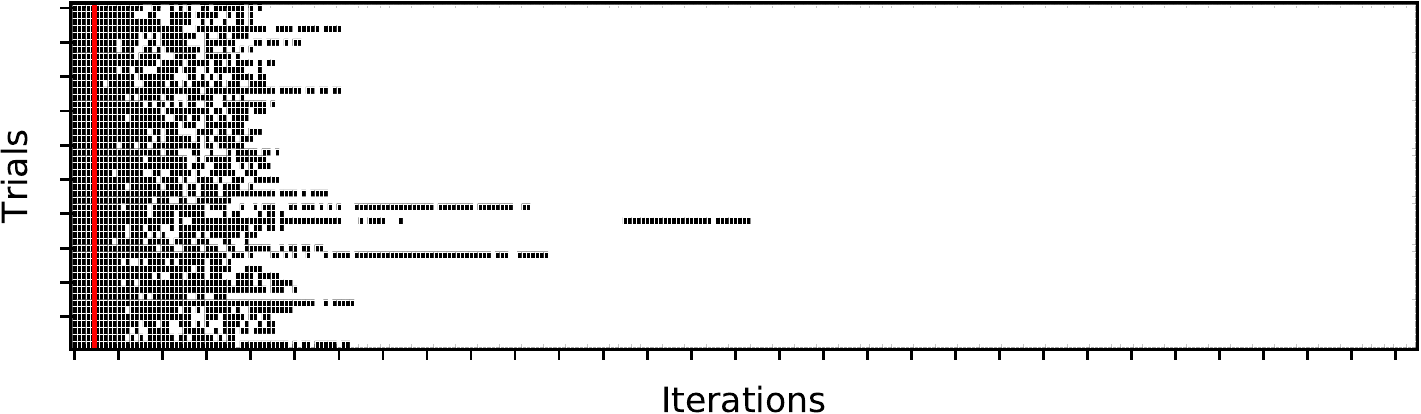} 
	\includegraphics[width=\imgW]{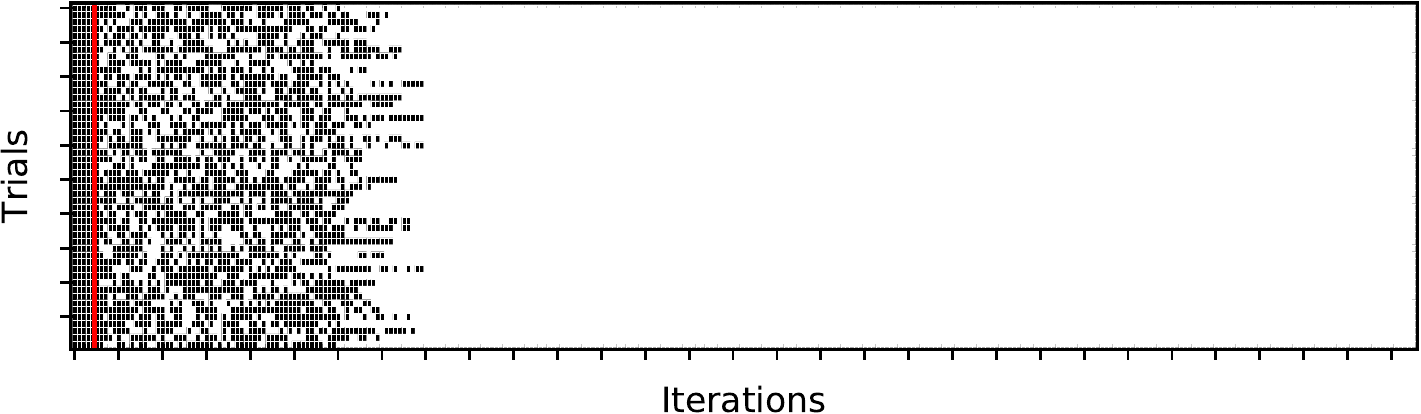}
	
	\includegraphics[width=\imgW]{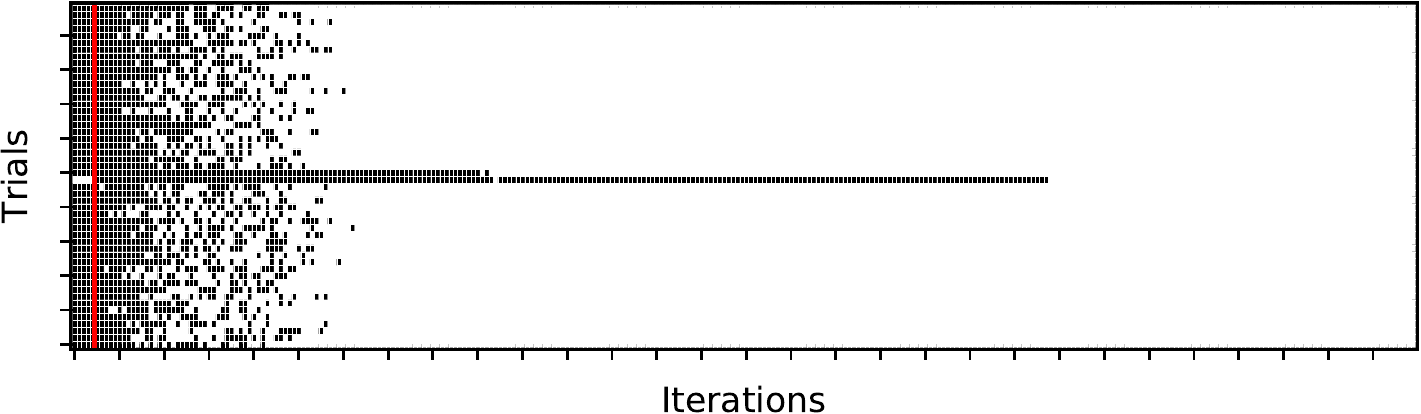} 
	\includegraphics[width=\imgW]{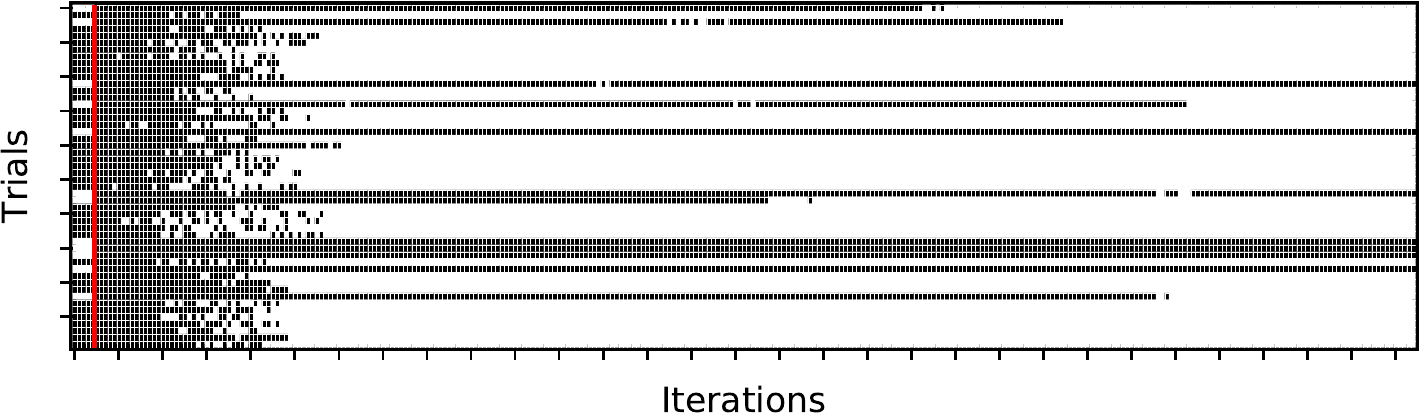} 
	\includegraphics[width=\imgW]{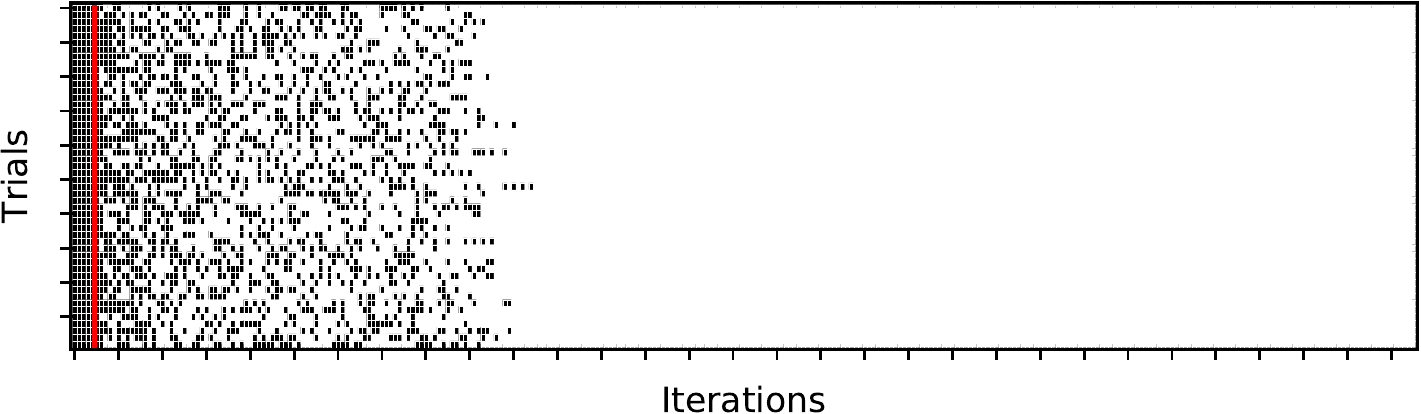}
	
	\caption{Improvements over iterations. Top to bottom: PSO, DFO and DE. Each black block represents finding an improvement to the previously found solution. Blank blocks on the left of the red vertical lines indicate failed trials. As illustrated, PSO exhibits the largest number of continuous, iteration-based improvements (albeit gradual). DFO and DE are shown to have less continuous improvements, with DFO presenting visible instances of escaping local minima for the first and second products.\label{fig:improvement}}
\end{figure}

In the experiments conducted here, iterations yielding an improvement over their preceding iteration are logged. Figure~\ref{fig:improvement} illustrates these improvements in the first $300$ iterations in $50$ independent trials for each of the algorithms when optimising the three products. It is shown that while PSO is less efficient than the other algorithms (as shown in Tables~\ref{tab:G-err}, \ref{tab:K-err} and \ref{tab:I-err}), it continuously improves on its current solution almost in every iteration until it terminates, either by reaching the optimum value, or getting trapped in a local minima. When optimising Imperial Black IPA, PSO shows iteration-based improvements in more than $97\%$ of the first $300$ iterations albeit failing in $4$~trials when optimising Guinness Extra Stout. DFO and DE fail in $1$~trial each and exhibit a comparable iteration-based improvement behaviour for this product\footnote{Note that the number of iterations allowed before termination for DE (in case of failing the trials) is less than PSO and DFO, as DE calls the fitness function twice in each iteration: one for evaluating the `target' vector ($\vec{x}$), and a second time to evaluate the mutated and crossed-over vector, the `trial' vector ($\vec{u}$). }.

The figures show that DFO exhibits a larger number of attempts to escape from potential local minima (where there are no solution improvements for several iterations, followed by repeated improvements as a result of escaping a potential local minima). This is visually evident in some trials for Guinness Extra Stout and Kozel Black. Escaping local minima could be a contributing factor in DFO's higher reliability (see eq.~\ref{eq:Swarm Reliability}), and therefore more optimal solution vectors which could be analysed for their diversity.

\subsection{Solution vectors diversity\label{subsec:solVecDiv}}

\begin{figure}[t]
	\centering
	PSO \hspace{0.24\textwidth} DFO \hspace{0.24\textwidth} DE \hspace{0.04\textwidth}
	
	\setlength{\imgW}{0.30\linewidth}
	
	\includegraphics[width=\imgW]{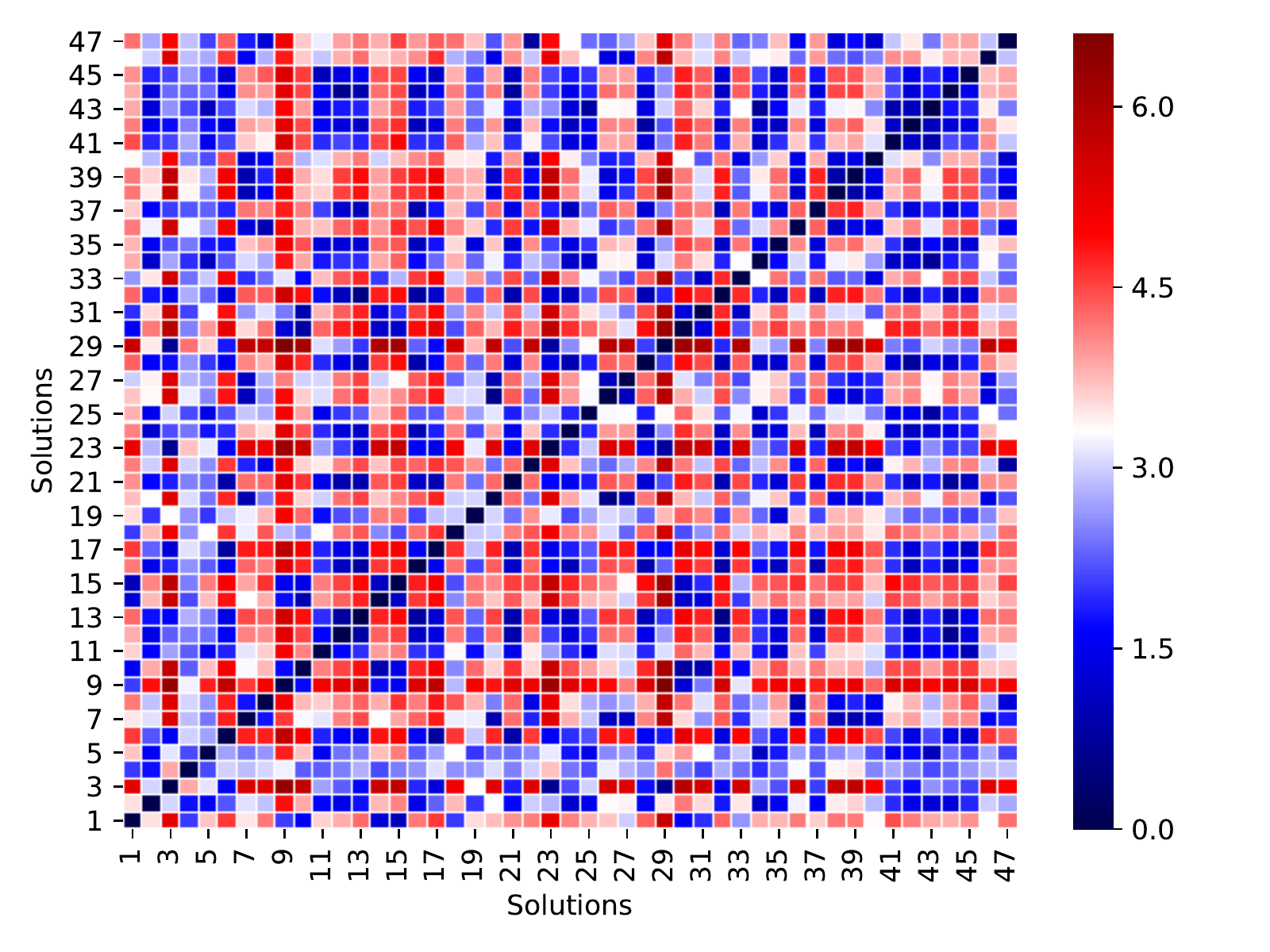} 
	\includegraphics[width=\imgW]{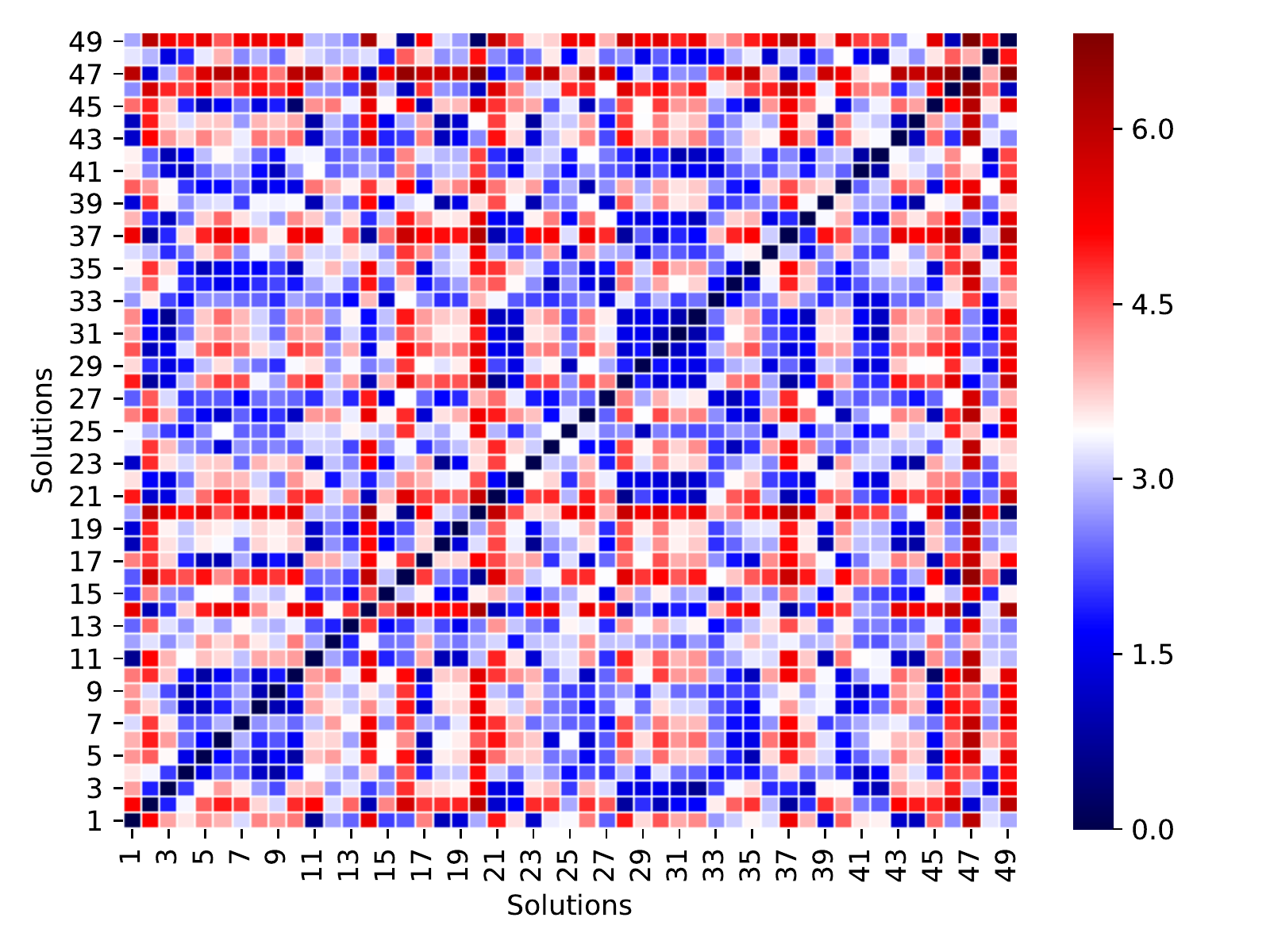} 
	\includegraphics[width=\imgW]{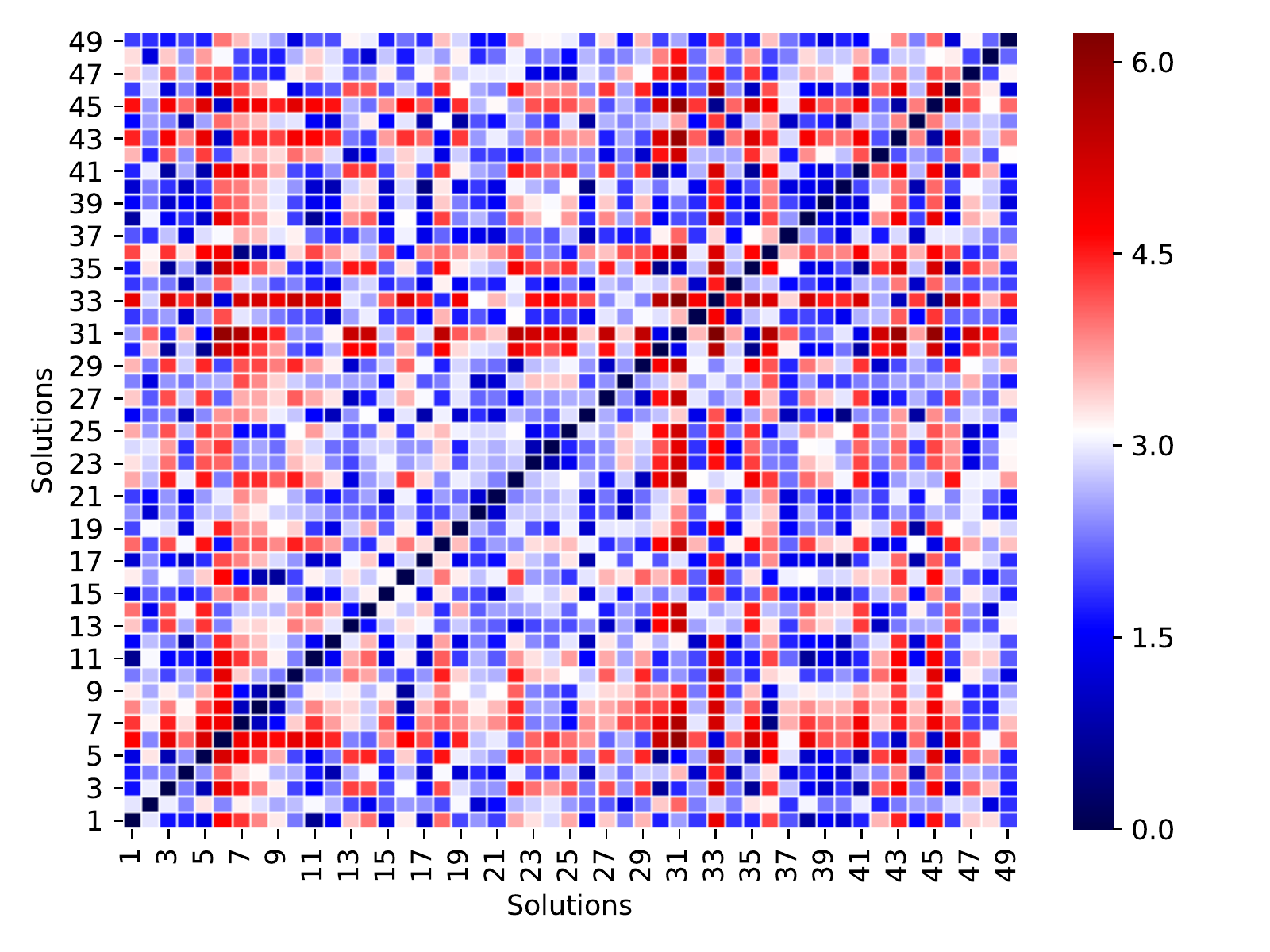} 
	
	\includegraphics[width=\imgW]{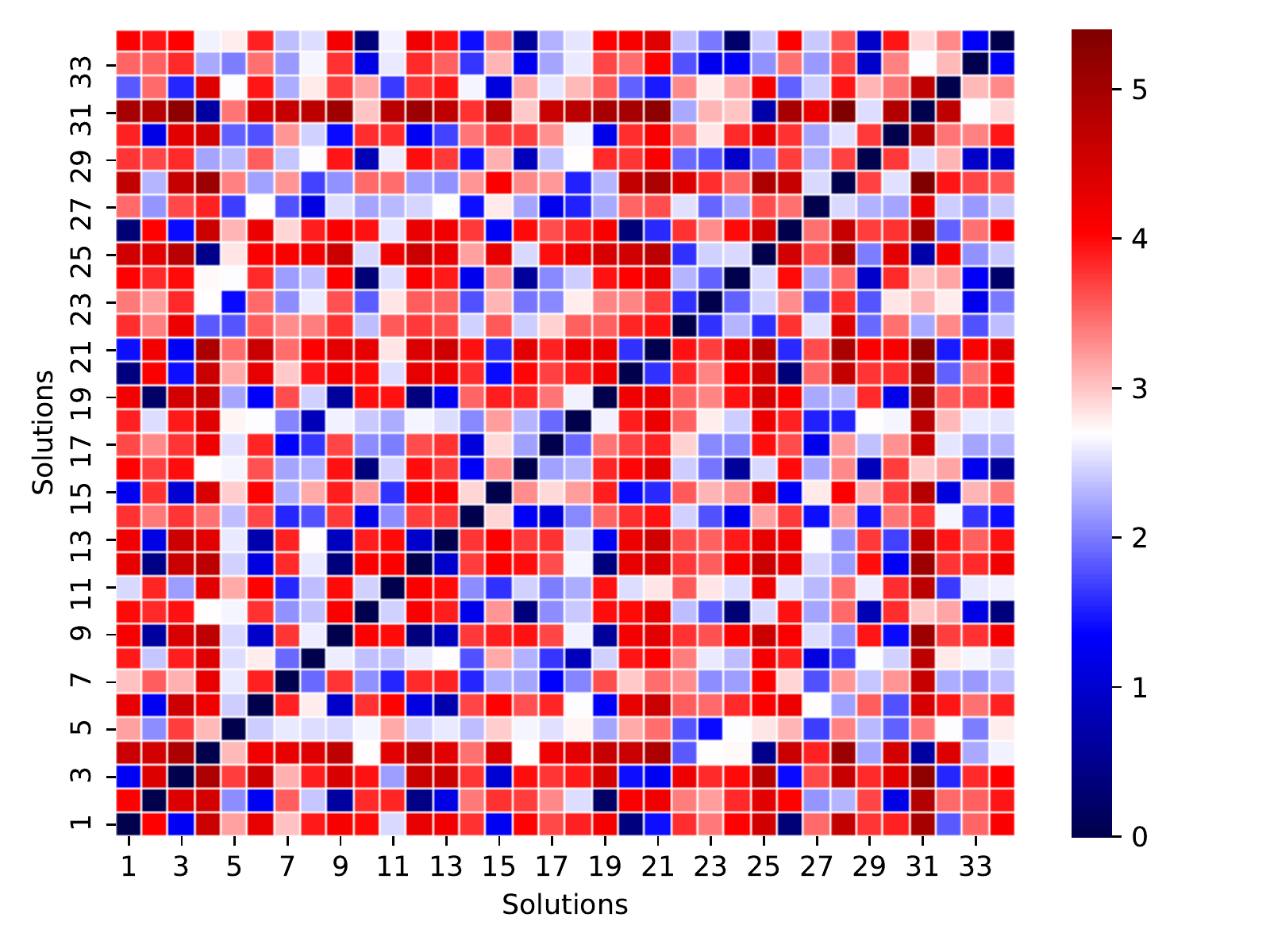} 
	\includegraphics[width=\imgW]{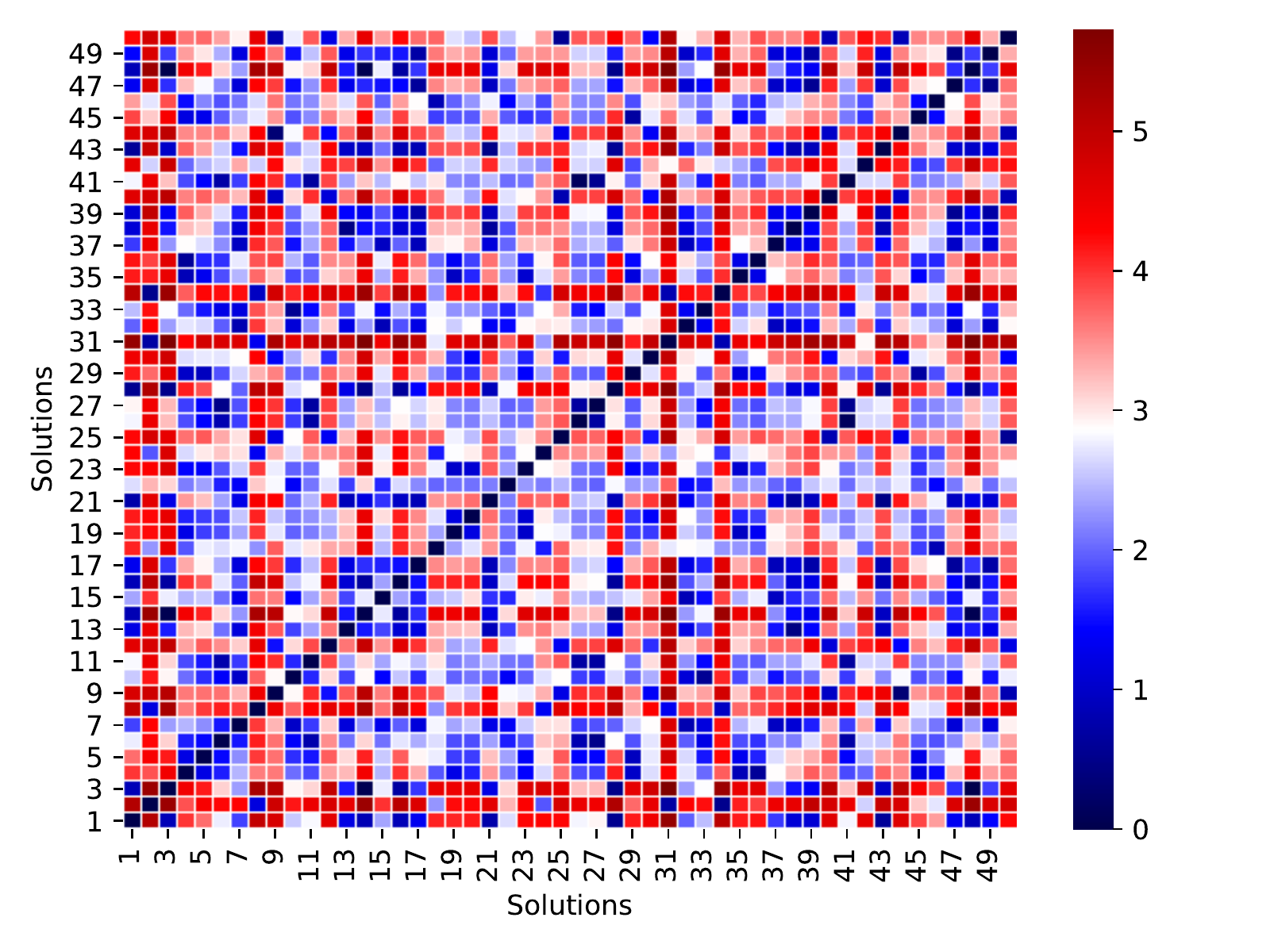} 
	\includegraphics[width=\imgW]{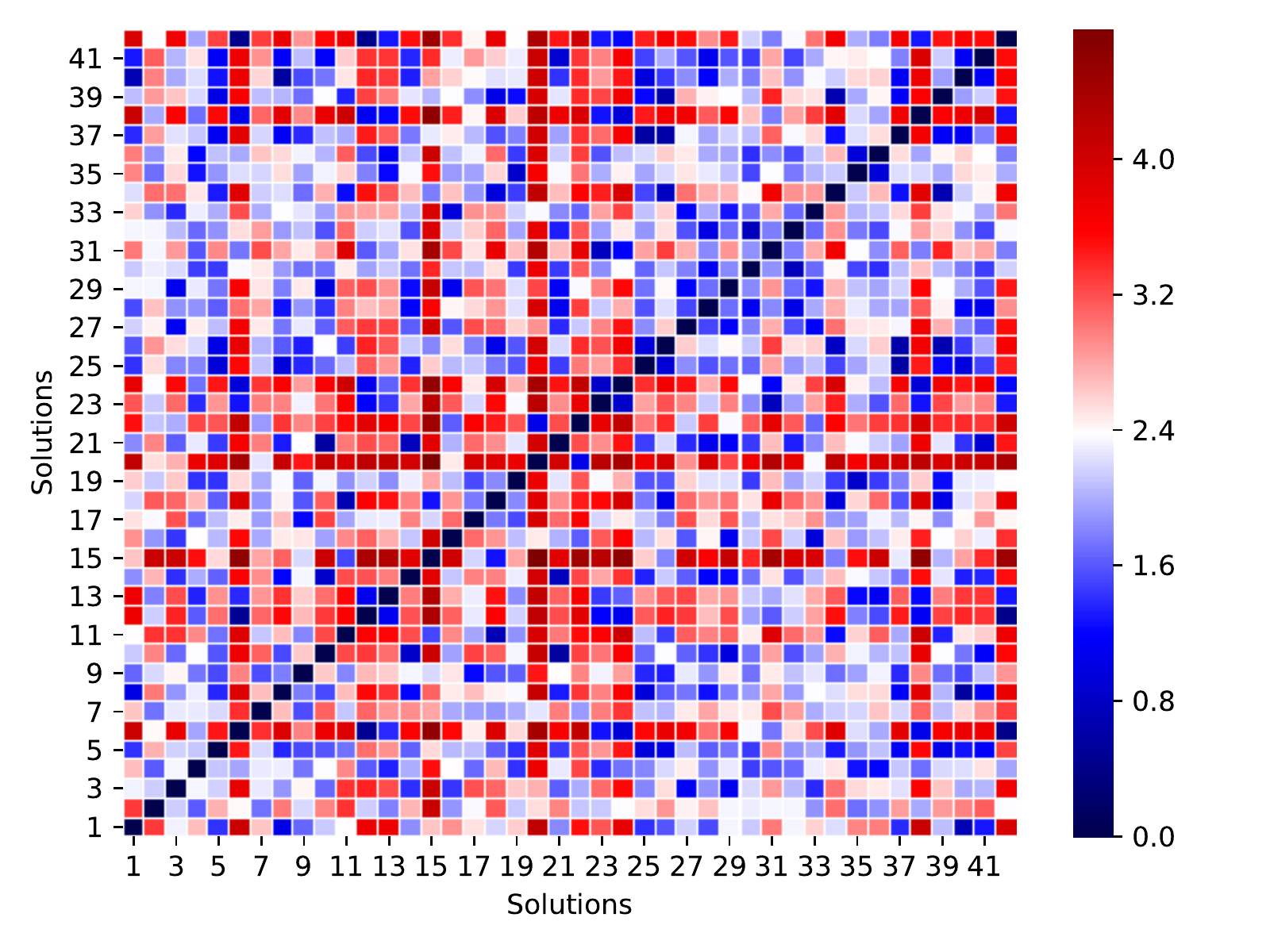} 
	
	\includegraphics[width=\imgW]{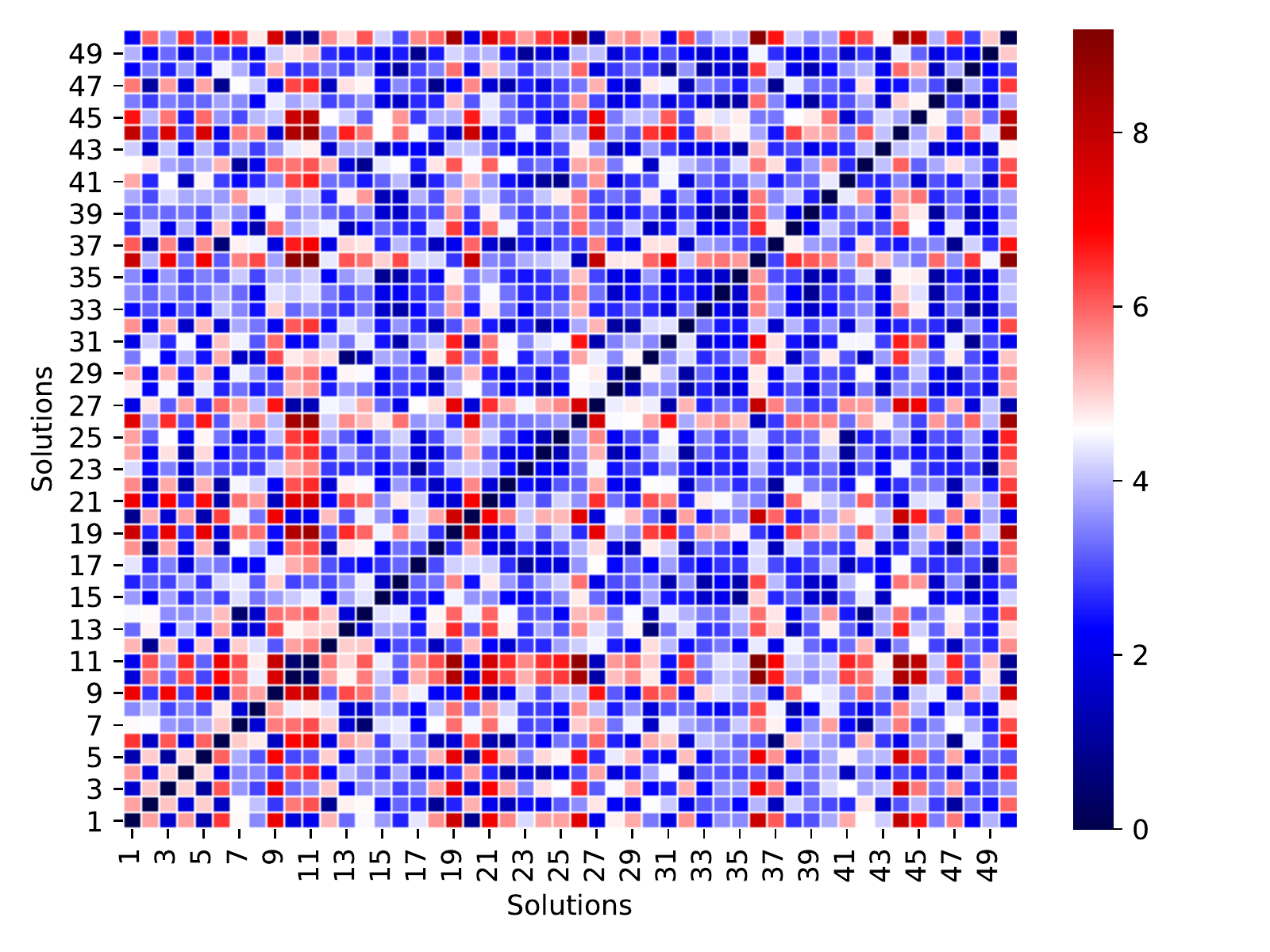} 
	\includegraphics[width=\imgW]{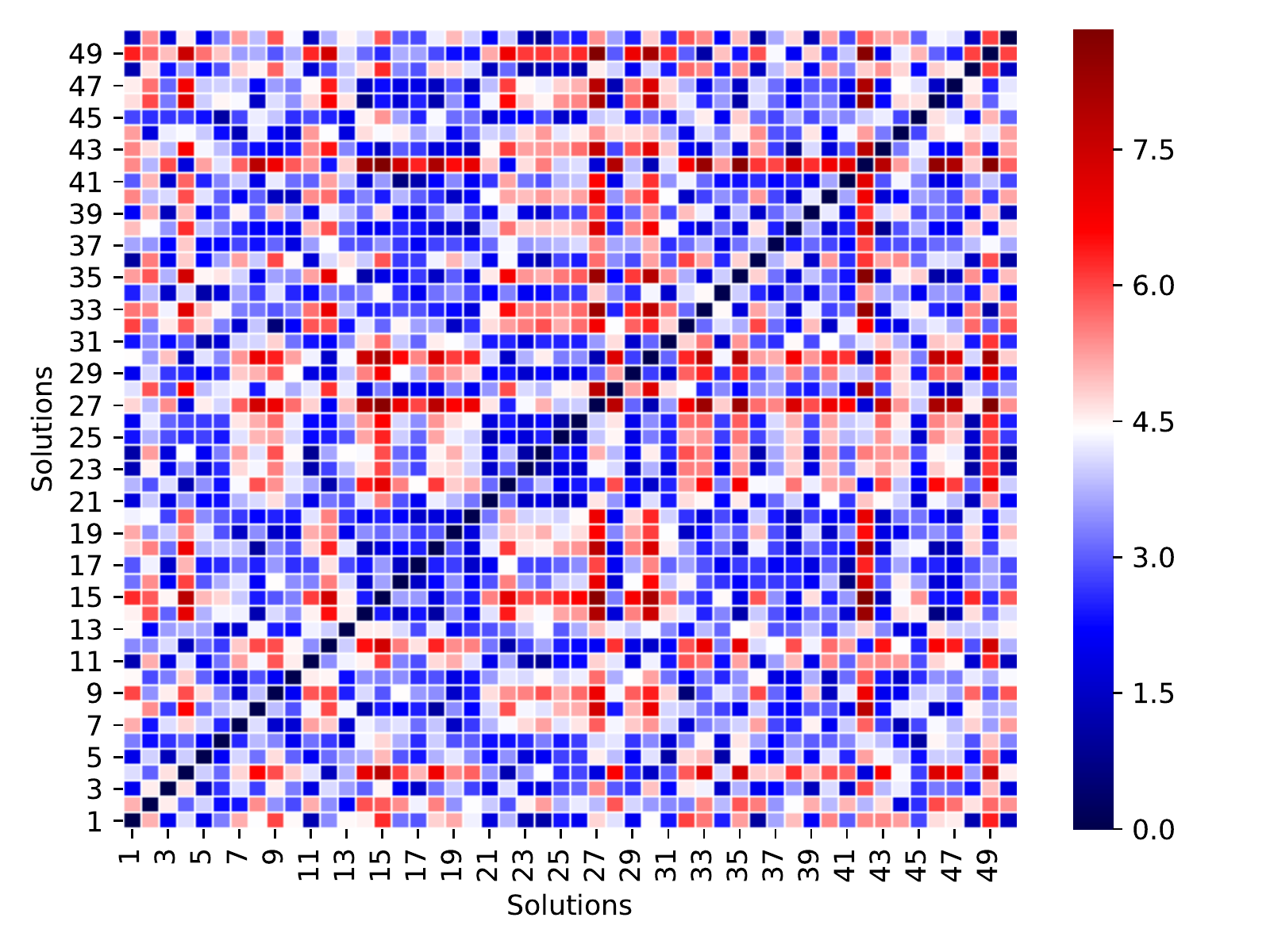} 
	\includegraphics[width=\imgW]{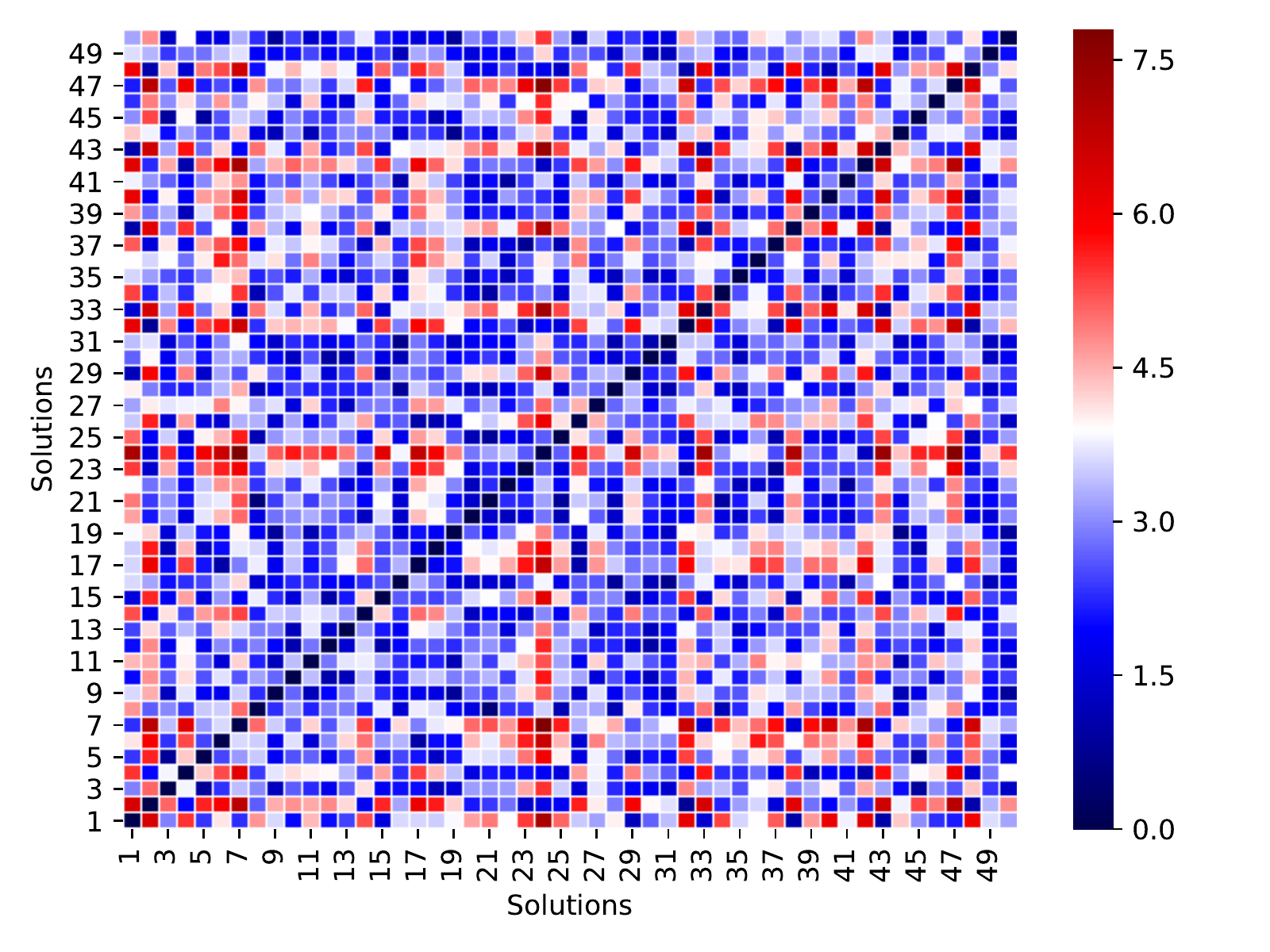} 
	
	\caption{Solution vector distances. Top: Guinness Extra Stout; middle: Kozel Black and bottom: Imperial Black IPA. Observing the heatmaps and the associated colour bars, on average, DFO generates the most varied solutions (on average) for products 1 and 3, and PSO for product 2. Table~\ref{tab:solDiv} presents the numerical summary of solution distance matrices. Taking each optimiser-product pair independently, DFO generates the most distant solutions for the first and second products and PSO for the third. Note the scales in the heatmaps are dependant on the solution distances (see the upper bounds of the colour bars); also only optimal solution vectors are included in this analysis (see Tables~\ref{tab:G-err}, \ref{tab:K-err}, \ref{tab:I-err}). \label{fig:disMat}}
\end{figure}

To evaluate the uniqueness of the already generated solution vectors, distances between each pair of solutions are studied. These values are presented as distance matrices in Figure~\ref{fig:disMat}. One of the practical implications of having `distant' solutions is their potentially radically different ingredient-combinations. In other words, in some extreme cases, some ingredients might be used entirely in one solution vector while remaining untouched in another. Making practically `unique' solutions available to the user allows them to choose based on their priorities or future process plans.
To numerically analyse the solution diversity for Guinness Extra Stout, Table~\ref{tab:solDiv}a shows that DFO presents the most distant solutions on average, followed by PSO; this is reaffirmed with the maximum distance found and is visually evident by comparing the upper bound of the colour bars in Figure~\ref{fig:disMat}-top, where the farthest pairs can be observed.
As for Black Kozel (Table~\ref{tab:solDiv}b), PSO exceeds in its average solutions diversity, followed by DFO, which however generates the two most distant solutions. In terms of the solution diversity for Imperial Black IPA (Table~\ref{tab:solDiv}c), on the contrary to the previous product and along the line of the first, DFO exhibits the largest on-average diversity in its solutions, however PSO produces the farthest two solutions. In all products, DE is shown to be producing solutions with the least distances.

\begin{figure}
	\setlength{\imgW}{0.30\linewidth}
	\centering
	PSO \hspace{0.3\textwidth} DFO \hspace{0.3\textwidth} DE 
	
	\includegraphics[width=\imgW]{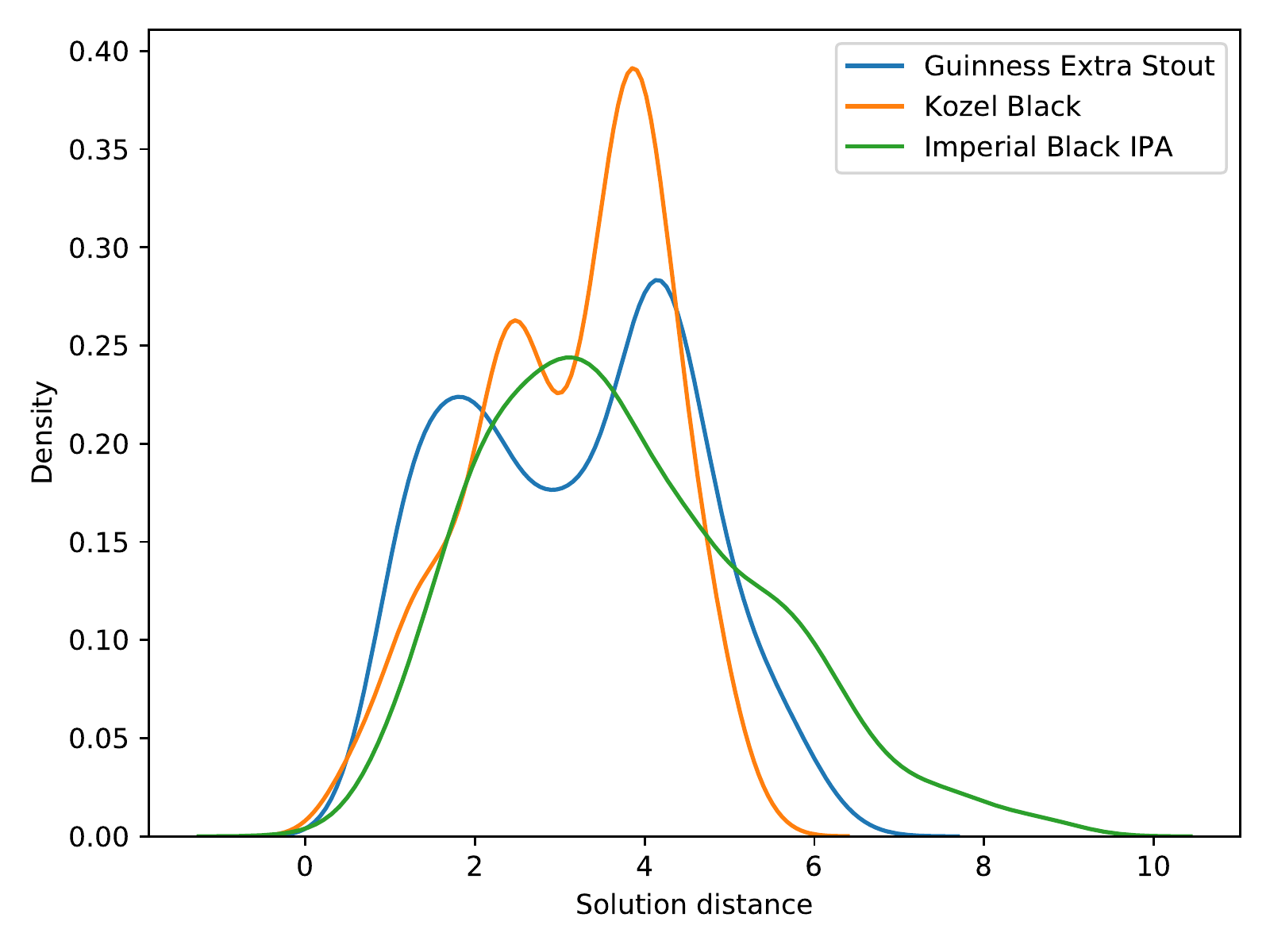} \hspace{5mm}
	\includegraphics[width=\imgW]{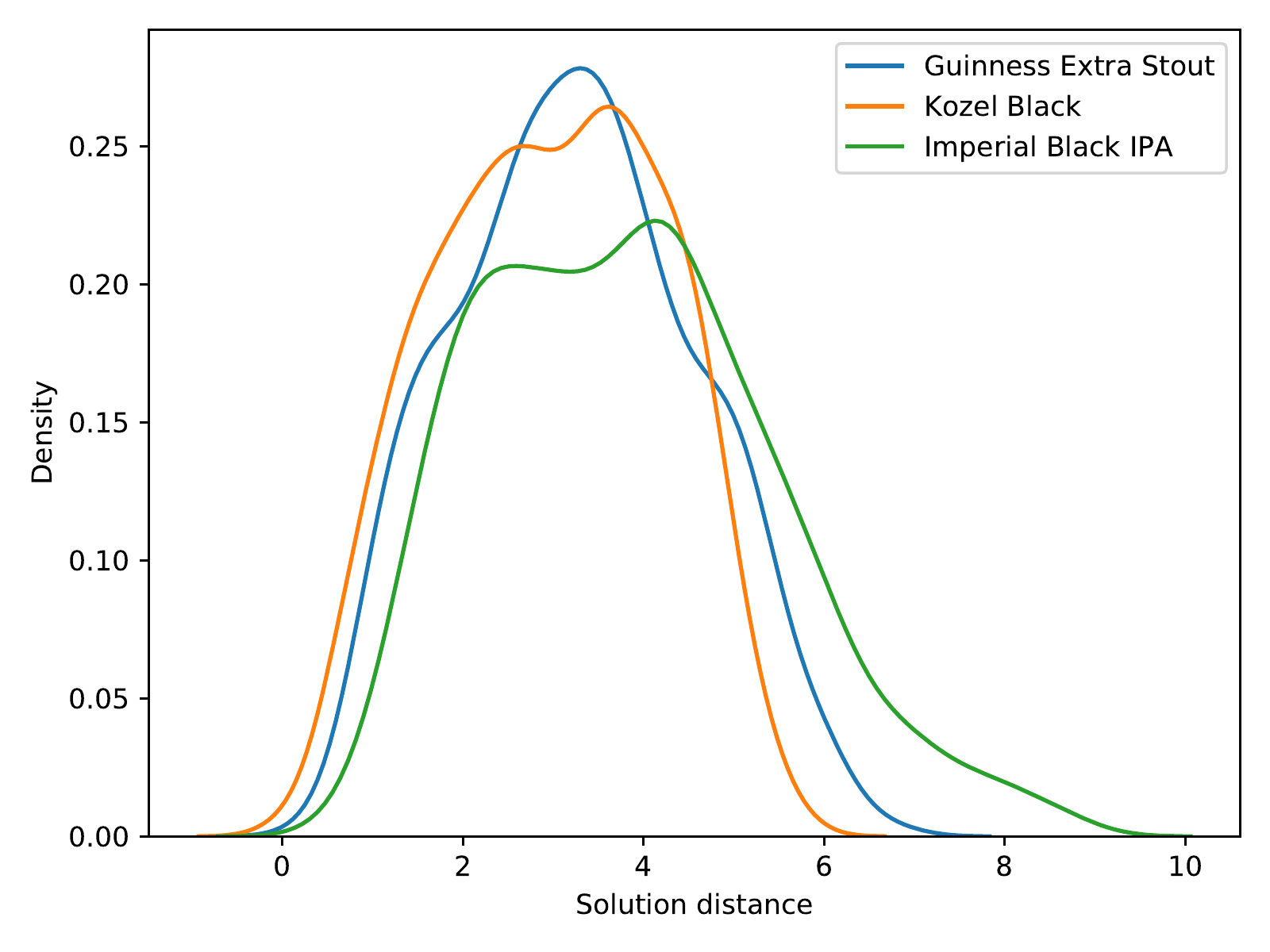} \hspace{5mm}
	\includegraphics[width=\imgW]{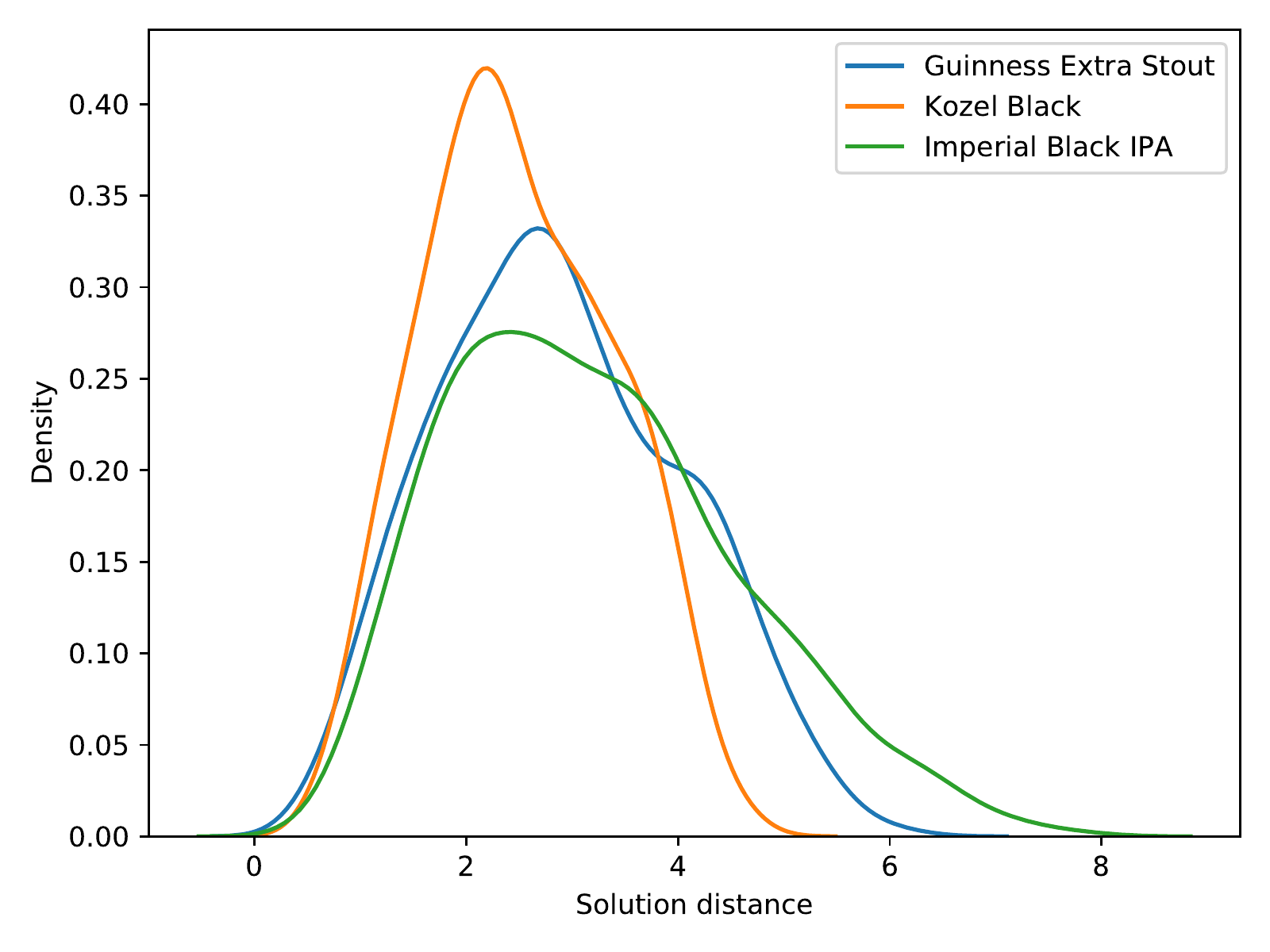} 
	
	\caption{Density of solution distances based on solution distance matrices in Figure~\ref{fig:disMat}.\label{fig:density}}
\end{figure}

In summary, DFO has generated the most distant solutions in the first two products, and the largest average distances in the first and third product. To further visualise the algorithms' behaviour, the density of the solutions distances (see Figure~\ref{fig:density}) shows that DFO is consistent in producing distant solutions in all three products.
The results in Figures~\ref{fig:disMat} and~\ref{fig:density} will be further discussed when solution clusters are presented.

\begin{table} [t]
	\centering \footnotesize
	\begin{tabular}{lccc}
		\toprule
		\textbf{(a) Guinness Extra Stout }& {PSO} & {DFO} & {DE}\tabularnewline
		\hline 
		{Mean} & 3.2620 & 3.3143 & 2.9441\tabularnewline
		{StDev} & 1.3330 & 1.2725 & 1.0936\tabularnewline
		{Min distance} & 0.5420 & 0.2506 & 0.4760\tabularnewline
		{Max distance} & 6.6126 & 6.8210 & 6.2258\tabularnewline
		{Farthest pair} & (9,29) & (20,47) & (31,33)\tabularnewline
		\bottomrule
	\end{tabular}
	
	\vspace{1mm}
	\begin{tabular}{lccc}
		\toprule 
		\textbf{(b) Kozel Black\;\;\;\;\;\;\;\;\;\;\;\;\;\;\;}& {PSO} & {DFO} & {DE}\tabularnewline
		\hline 
		{Mean} & 3.1823 & 3.0192 & 2.5479\tabularnewline
		{StDev} & 1.1757 & 1.1734 & 0.8733\tabularnewline
		{Min distance} & 0.1997 & 0.0201 & 0.4158\tabularnewline
		{Max distance} & 5.3978 & 5.7259 & 4.7637\tabularnewline
		{Farthest pair} & (28,31) & (31,48) & (15,20)\tabularnewline
		\bottomrule 
	\end{tabular}
	
	\vspace{1mm}
	\begin{tabular}{lccc}
		\toprule 
		\textbf{(c) Imperial Black IPA\;\;\;}& {PSO} & {DFO} & {DE}\tabularnewline
		\hline 
		{Mean} & 3.8022 & 3.9033 & 3.2707\tabularnewline
		{StDev} & 1.5346 & 1.5551 & 1.2979\tabularnewline
		{Min distance} & 0.4379 & 0.5221 & 0.5176\tabularnewline
		{Max distance} & 9.1789 & 8.8227 & 7.7959\tabularnewline
		{Farthest pair} & (11,36) & (15,42) & (7,24)\tabularnewline
		\bottomrule 
	\end{tabular}
	\caption{Solutions diversity \label{tab:solDiv}}
\end{table}

\subsection{Solution clustering\label{subsec:solClustering}}

\begin{figure}[t]
	\setlength{ \imgW }{.25\linewidth}
	\centering
	
	Guinness Extra Stout \hspace{0.1\textwidth} Kozel Black \hspace{0.1\textwidth} Imperial Black IPA 
	
	\fbox{\includegraphics[width=\imgW]{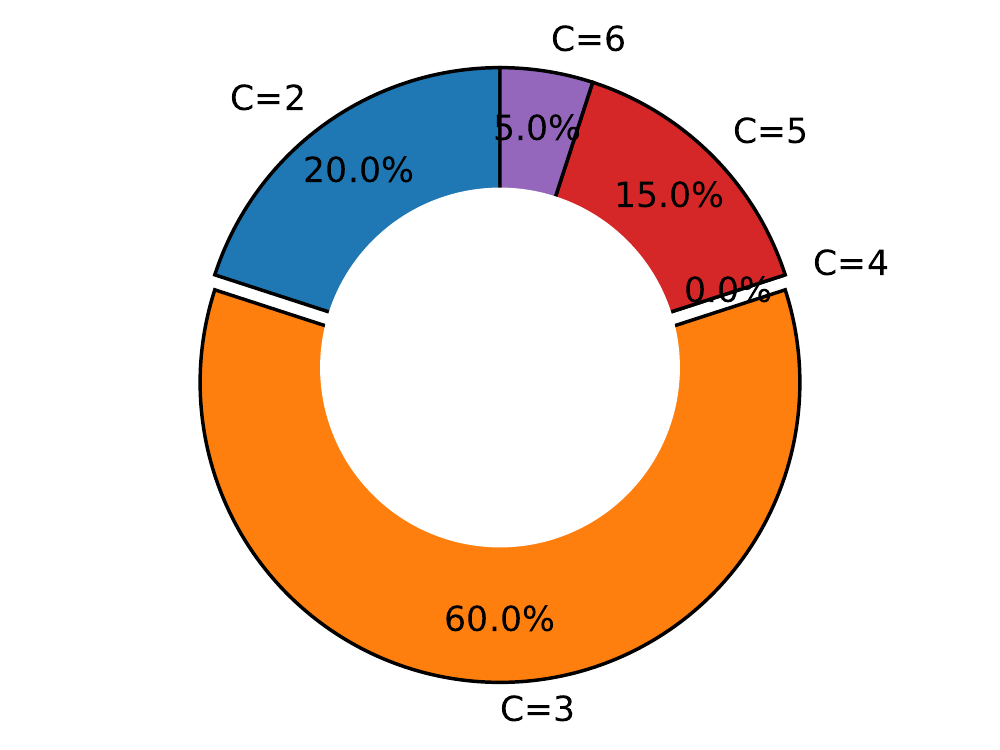}}
	\fbox{\includegraphics[width=\imgW]{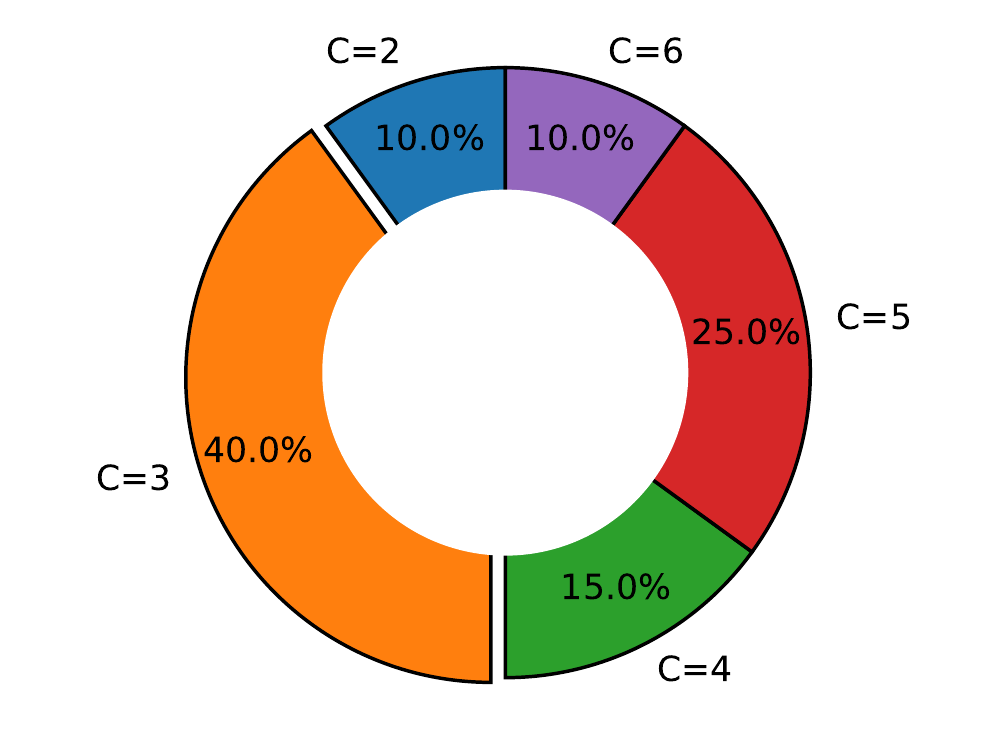}}
	\fbox{\includegraphics[width=\imgW]{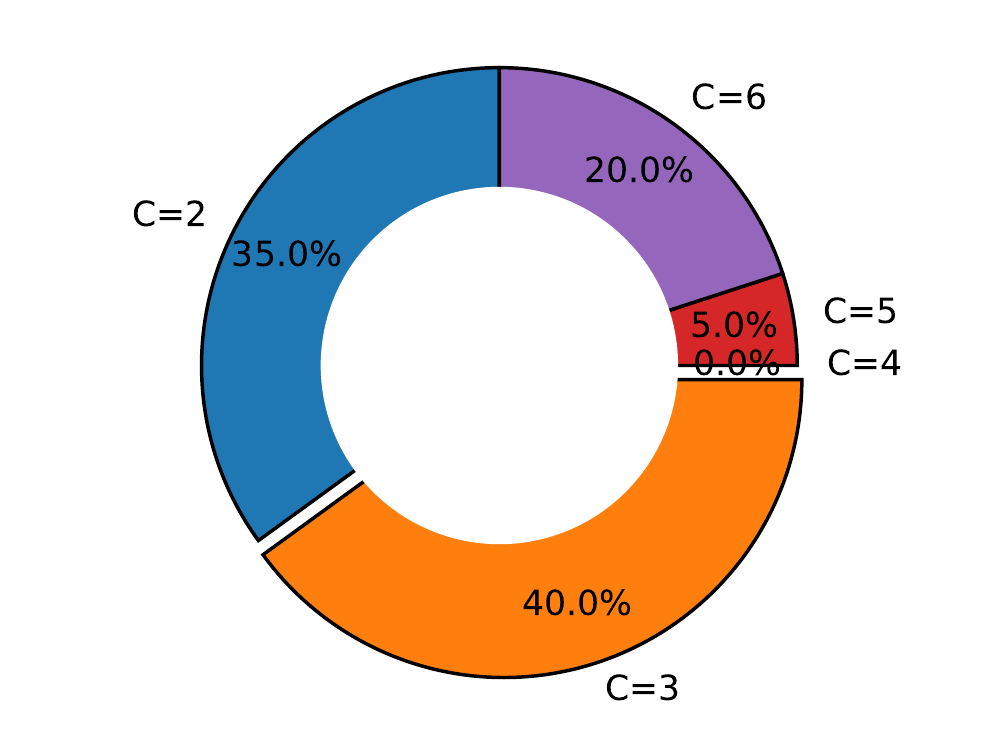}}
	
	\fbox{\includegraphics[width=\imgW]{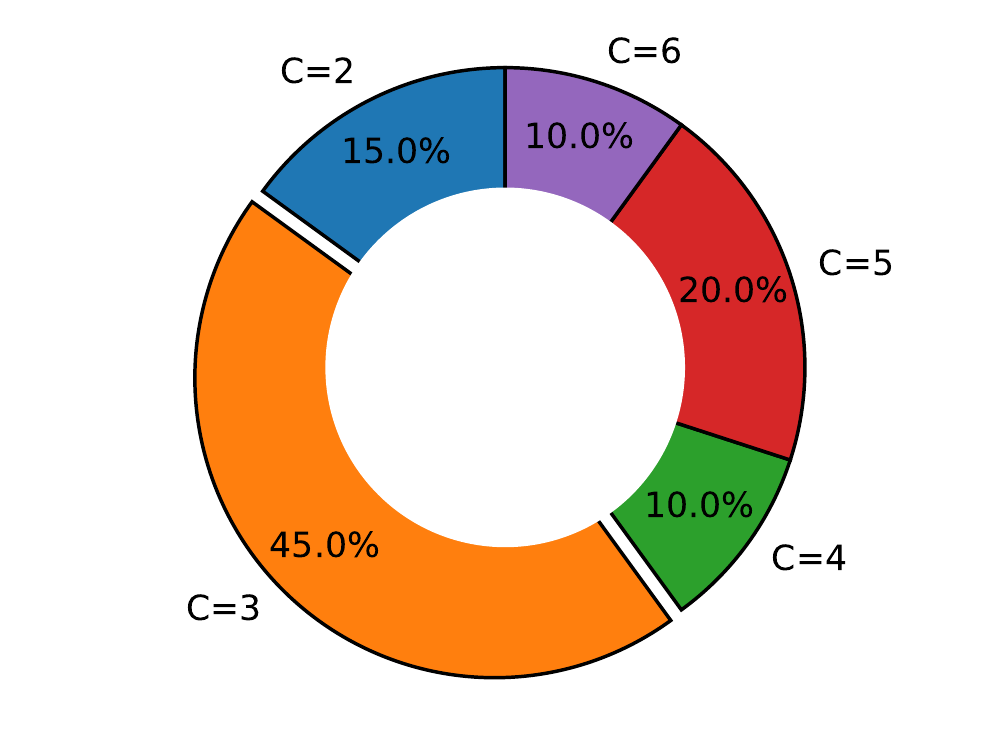}}
	\fbox{\includegraphics[width=\imgW]{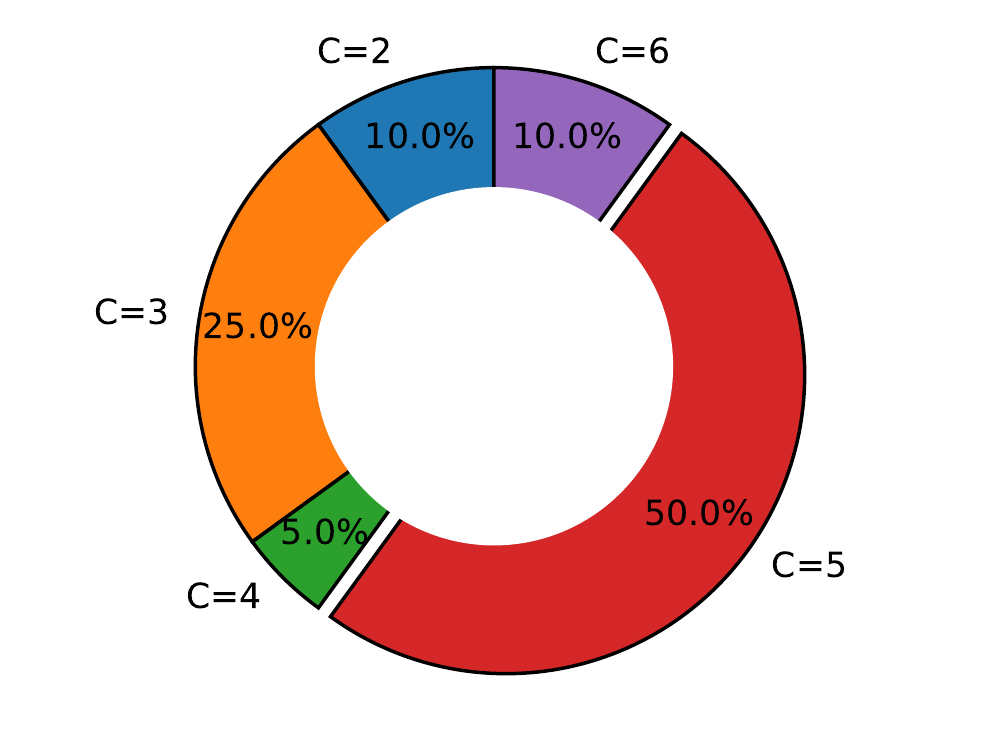}}
	\fbox{\includegraphics[width=\imgW]{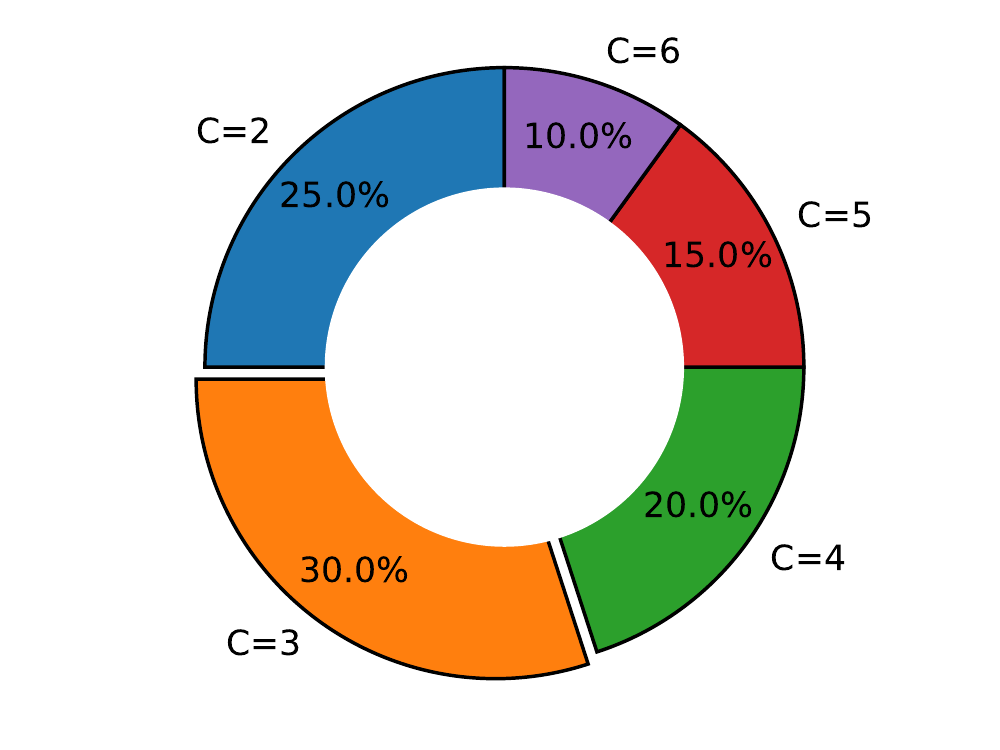}}
	
	\fbox{\includegraphics[width=\imgW]{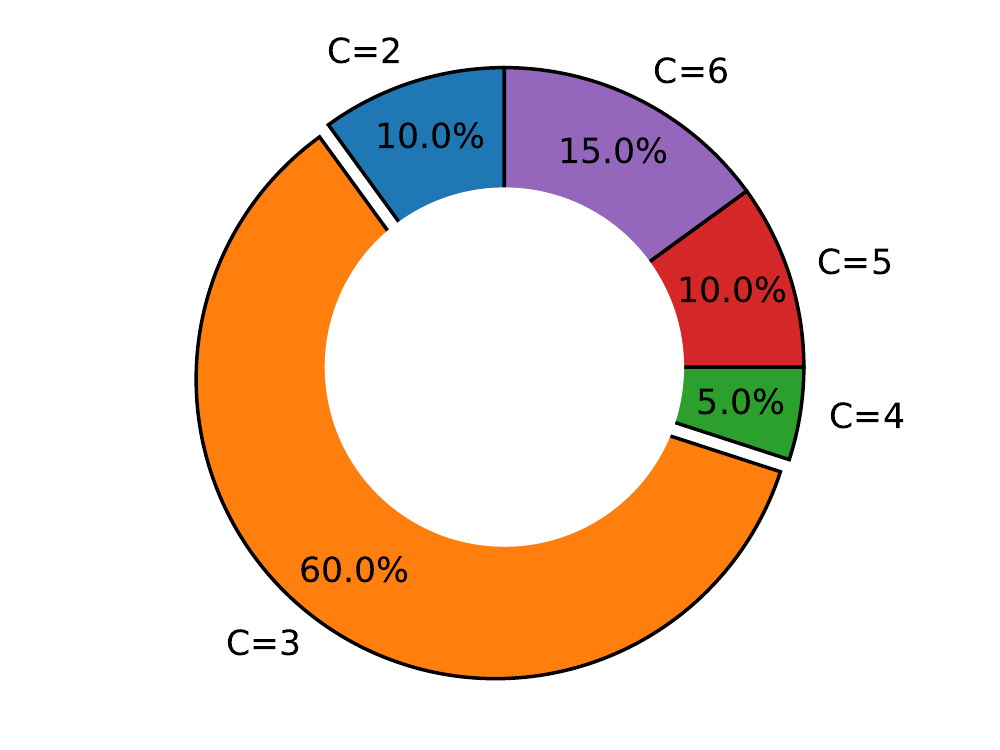}}
	\fbox{\includegraphics[width=\imgW]{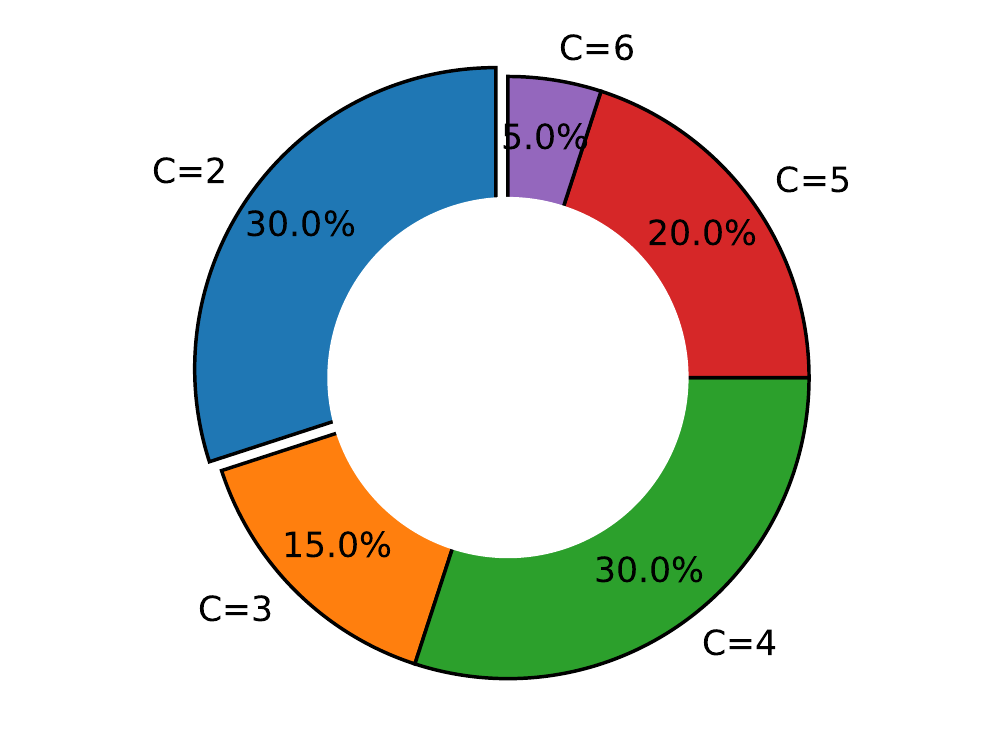}}
	\fbox{\includegraphics[width=\imgW]{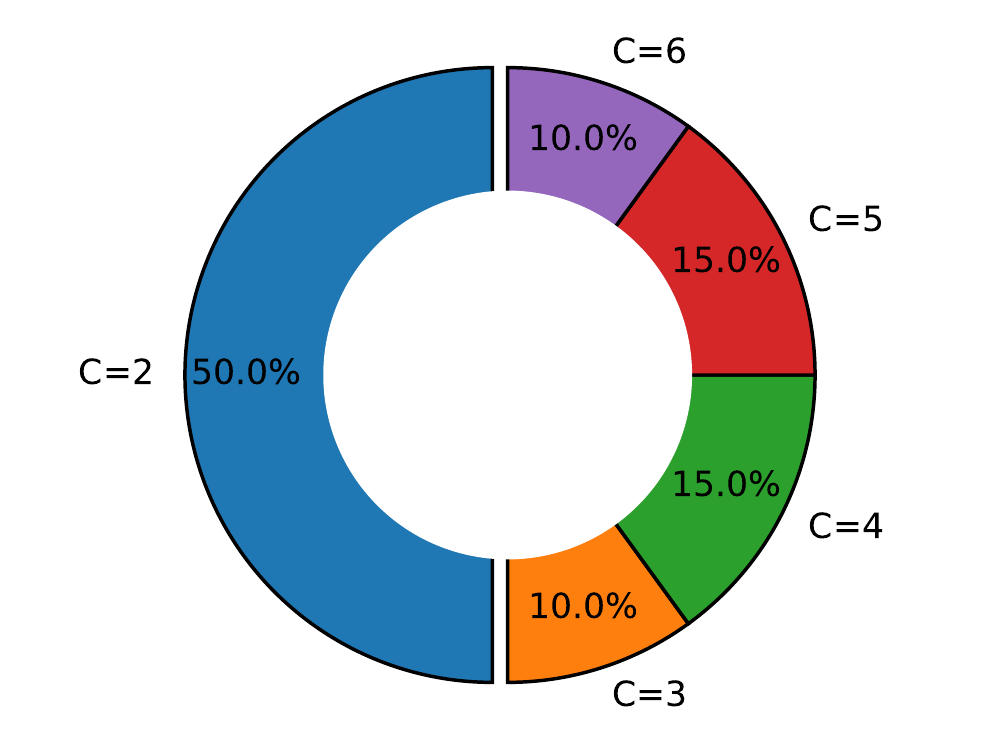}}
	
	\caption{Number of clusters. From top to bottom: PSO, DFO and DE. For each optimiser-product pair, $C=\{2,...,6\}$ represents the number of clusters from $2$ to $6$, whose strength proportion is determined by taking into account $20$ clustering indices.\label{fig:clusters}}
\end{figure}

In order to further limit the distinct classes of solutions based on their propinquity, clustering is applied, and therefore the challenge of selecting `unique' solutions which are farther apart is reduced. This grants further freedom to the user in adhering to their other production priorities.
To identify distinct clusters, K-means~\cite{macqueen1967some} is utilised, and to find the best number of clusters for each of the cloned beer, twenty indices~\cite{charrad2012nbclust} (e.g. \cite{CH_calinski1974dendrite,KL_krzanowski1988criterion,Cindex_hubert1976general,Hartigan1975clustering,DB_davies1979cluster,Silhouettes_rousseeuw1987silhouettes,Duda1973pattern}) are used. The majority rule is then applied to find the best number of clusters as shown in Figure~\ref{fig:clusters}.

As presented in Table~\ref{tab:clustering}a, when clustering the first product, Guinness Extra Stout, the most evenly distributed clusters are created by DFO, however with the least majority, while PSO and DE have a higher majority at the expense of returning an imbalanced number of solutions in each cluster. This can be explained by observing the density of solution distances for the first product, where some solution distances are more dense than others (see Figure~\ref{fig:density}) . When clustering the solutions for Black Kozel, DFO produces $5$ clusters, the maximum number of clusters, and the highest majority among the optimisers. This can be explained as the density of the solution distances for this product is the widest for DFO; on the contrary, the narrowest solution distance density for this product belongs to DE which returns two clusters\footnote{Note that when there is a tie in the number of clusters, as it is for DE optimising Kozel, it is recommended to choose the lower number.}. Although with the highest majority and the most evenly distribution solutions, the same is applicable for Imperial Black IPA and DE.

Further to the uniqueness of individual solutions themselves, at least two distinct ones from each set of optimising tasks can be selected (one from each cluster). Additionally, distance thresholds between clusters can be analysed by using methods such as hierarchical or agglomerative clustering approaches, which is a topic for ongoing research.

\begin{table} 	
	\centering \footnotesize
	\begin{tabular*}{\columnwidth}{@{\extracolsep{\fill}} lrrrr}
		\toprule 
		\textbf{(a) Guinness Extra Stout }& Cluster 1 & Cluster 2 & Cluster 3 & Majority\tabularnewline
		\hline 
		PSO & 25 (53\%) & 9 (19\%) & 13 (28\%) & 12 (60\%)\tabularnewline
		DFO & 18 (37\%) & 15 (31\%) & 16 (33\%) & 9 (45\%)\tabularnewline
		DE & 10 (20\%) & 12 (24\%) & 27 (55\%) & 12 (60\%)\tabularnewline
		\bottomrule 
	\end{tabular*}
	
	\vspace{3mm} 
	
	\begin{tabular*}{\columnwidth}{@{\extracolsep{\fill}} lrrrrrr}
		\toprule 
		\textbf{(b) Kozel Black}& Cluster 1 & Cluster 2 & Cluster 3 & Cluster 4 & Cluster 5 & Majority\tabularnewline
		\hline 
		PSO & 15 (44\%) & 11 (32\%) & 8 (24\%) & -- & -- & 8 (40\%)\tabularnewline
		DFO & 11 (22\%) & 7 (14\%) & 10 (20\%) & 16 (32\%) & 6 (12\%) & 10 (50\%)\tabularnewline
		DE & 13 (31\%) & 29 (69\%) & -- & -- & -- & 6 (30\%)\tabularnewline
		\bottomrule 
	\end{tabular*}	
	
	\vspace{3mm} 
	
	\begin{tabular*}{\columnwidth}{@{\extracolsep{\fill}} lrrrr}
		\toprule 
		\textbf{(c) Imperial Black IPA} & Cluster 1 & Cluster 2 & Cluster 3 & Majority\tabularnewline
		\hline 
		PSO & 13 (26\%) & 19 (38\%) & 18 (36\%) & 8 (40\%)\tabularnewline
		DFO & 25 (50\%) & 7 (14\%) & 18 (36\%) & 6 (30\%)\tabularnewline
		DE & 24 (48\%) & 26 (52\%) & -- & 10 (50\%)\tabularnewline
		\bottomrule 
	\end{tabular*}	
	\caption{Solution clusters\label{tab:clustering}}
\end{table}

\section{Conclusion}
\label{sec:conclusion}
The high experimental costs associated with the beer brewing process is shown to be efficiently reducible by taking into account the organoleptic characteristics along with the in-stock inventory. 
In this work, three swarm intelligence and evolutionary computation techniques are presented to automate the quantitative \textit{ingredients selection}, which is one of the key experimental aspects of brewing, specially in low cost production environments.

In terms of the performance measures, DFO is shown to be the most accurate and reliable algorithm, as well as the most efficient optimiser with statistically significant outperformance when compared to the other algorithms, followed by DE (Table~\ref{tab:summary-err-eff-rel}). 
Studying the iteration-based improvement, PSO is shown to present persistent improvement, with DFO exhibiting several cases of escaping local minima, which could be a contributing factor to its higher reliability (Figure~\ref{fig:improvement}). 
Analysing solution vectors diversity, DFO, on average, has produced the most distant solutions for two of the products, followed by PSO (Table~\ref{tab:solDiv}). 
To further analyse the distinctness of the optimisers' solutions for each product, the optimal number of clusters are derived by the majority rule with $20$ clustering indices. The algorithms are shown to be capable of producing diverse set of solutions, with DFO producing solutions with the same or more clusters in the aforementioned products (Table~\ref{tab:clustering}).

The presented approach alleviates the challenges of generating new and \textit{dynamically changing} recipes based on their organoleptic properties. This is an attractive feature for both commercial producers where varieties and quantities of ingredients are not hard constraints; and, in less equipped setups, with stronger ingredients-based constraints, allowing the design of high quality beer.

As part of ongoing and future work, in addition to investigating other case studies and alternative inventories, we are exploring the ability of the algorithms to adjust to changes to organoleptic characteristics of beers during the optimisation process, therefore, studying the impact of the population diversity further. 
Another topic for future research is the use of multi-objective optimisers and investigating how the reported results can be used to improve their performance in the context of the problem discussed.
Additionally, we are adding the more complex \textit{flavour} and \textit{aroma} profiles as well as \textit{foam} characteristics, which are dependent, among others, on the fermentables and hops. Furthermore, each hop's boiling time could be added as an extra dimension which would impact the aforementioned aroma and flavour profiles of the result.

\section*{Acknowledgement}
The authors would like to thank  Edmund Oetgen for taking the initial steps of the implementation, and Christian Juri for the real-world trial of the `swarm beer system' in form of the Indian Pale Ale, \textit{FLIPA}, with pleasantly memorable results!

\bibliographystyle{plain}
\bibliography{../../myRef}

\end{document}